\documentclass[final,journal,letterpaper,twocolumn]{IEEEtran}
\newcommand{\lia}[1]{\textcolor{black}{#1}}
\newcommand{\lib}[1]{\textcolor{black}{#1}}
\newcommand{\lic}[1]{\textcolor{black}{#1}}
\newcommand{\lid}[1]{\textcolor{blue}{#1}}
\ifCLASSINFOpdf
  \usepackage[pdftex]{graphicx}
  \graphicspath{{./figures/}}
  \DeclareGraphicsExtensions{.eps,.pdf,.jpeg,.png,.tif}
\else
\fi
\usepackage{hyperref}

\usepackage{amsmath}

\usepackage{algorithm,algorithmic}

\usepackage{booktabs}

\usepackage{xcolor}

\usepackage{array}

\usepackage{amsfonts}
\usepackage{bm}
\usepackage{graphicx}
\usepackage{booktabs}
\usepackage{xfrac}
\usepackage{threeparttable}
\usepackage{multirow}
\usepackage{multicol}
\usepackage{rotating}
\usepackage{changepage}
\usepackage{lineno,hyperref}
\modulolinenumbers[5]
\usepackage{makecell}
\newcolumntype{P}[1]{>{\centering\arraybackslash}p{#1}}

\ifCLASSOPTIONcompsoc
  \usepackage[caption=false,font=normalsize,labelfont=sf,textfont=sf]{subfig}
\else
  \usepackage[caption=false,font=footnotesize]{subfig}
\fi

\begin{document}

\title{Semi-Supervised Building Footprint Generation with Feature and Output Consistency Training }

\author{Qingyu Li,~\IEEEmembership{Student Member,~IEEE},
        Yilei Shi,~\IEEEmembership{Member,~IEEE},
        and~Xiao Xiang~Zhu,~\IEEEmembership{Fellow,~IEEE}
\thanks{This work is supported by the European Research Council (ERC) under the European Union's Horizon 2020 research and innovation programme (grant agreement No. [ERC-2016-StG-714087], Acronym: \textit{So2Sat}), by the Helmholtz Association through the Framework of Helmholtz AI (grant  number:  ZT-I-PF-5-01) - Local Unit ``Munich Unit @Aeronautics, Space and Transport (MASTr)'' and Helmholtz Excellent Professorship ``Data Science in Earth Observation - Big Data Fusion for Urban Research''(grant number: W2-W3-100), by the German Federal Ministry of Education and Research (BMBF) in the framework of the international future AI lab "AI4EO -- Artificial Intelligence for Earth Observation: Reasoning, Uncertainties, Ethics and Beyond" (grant number: 01DD20001) and by German Federal Ministry of Economics and Technology in the framework of the "national center of excellence ML4Earth" (grant number: 50EE2201C). This work is also part of the project "Investigation of building cases using AI" funded by Bavarian State Ministry of Finance and Regional Identity (StMFH) and the Bavarian Agency for Digitization, High-Speed Internet and Surveying.}
\thanks{\emph{Corresponding author: Xiao Xiang Zhu.}}

\thanks{Q.~Li, and X.X.~Zhu are with the Chair of Data Science in Earth Observation, Technische Universit{\"a}t M{\"u}nchen (TUM), 80333 Munich, Germany and the Remote Sensing Technology Institute (IMF), German Aerospace Center (DLR), 82234 We\ss ling, Germany (e-mails: qingyu.li@tum.de; xiaoxiang.zhu@dlr.de)}

\thanks{Y.~Shi is with the Chair of Remote Sensing Technology, Technische Universit{\"a}t M{\"u}nchen (TUM), 80333 Munich, Germany (e-mail: yilei.shi@tum.de)}

}

%
%

\markboth{submitted to IEEE TRANSACTIONS ON GEOSCIENCE AND REMOTE SENSING, 2021}
{A. B \MakeLowercase{\textit{et al.}}:A Boundary Regularization Network for the Building Footprint Generation from Remote Sensing Satellite imagery}
%

\maketitle

\begin{abstract}
\lid{This is the preprint version, to read the final version
please go to IEEE Transactions on Geoscience and Remote
Sensing on IEEE Xplore.}Accurate and reliable building footprint maps are vital to urban planning and monitoring, and most existing approaches fall back on convolutional neural networks (CNNs) for building footprint generation. However, one limitation of these methods is that they require strong supervisory information from massive annotated samples for network learning. State-of-the-art semi-supervised semantic segmentation networks with consistency training can help to deal with this issue by leveraging a large amount of unlabeled data, which encourages the consistency of model output on data perturbation. Considering that rich information is also encoded in feature maps, we propose to integrate the consistency of both features and outputs in the end-to-end network training of unlabeled samples, enabling to impose additional constraints. Prior semi-supervised semantic segmentation networks have established the cluster assumption, in which the decision boundary should lie in the vicinity of low sample density. In this work, we observe that for building footprint generation, the low-density regions are more apparent at the intermediate feature representations within the encoder than the encoder's input or output. Therefore, we propose an instruction to assign the perturbation to the intermediate feature representations within the encoder, which considers the spatial resolution of input remote sensing imagery and the mean size of individual buildings in the study area. The proposed method is evaluated on three datasets with different resolutions: Planet dataset (3 m/pixel), Massachusetts dataset (1 m/pixel), and Inria dataset (0.3 m/pixel). Experimental results show that the proposed approach can well extract more complete building structures and alleviate omission errors.
\end{abstract}


\section{Introduction}
Building footprint generation is a hot topic in the community of remote sensing, which involves numerous applications such as identifying undocumented buildings and assessing building damage after natural disasters. Remote sensing imagery that offers potential for meaningful geospatial target extraction \lia{on} a large scale, becomes a fundamental data source for building footprint generation. However, obtaining accurate and reliable building footprint maps from remote sensing imagery is still challenging due to several reasons. On the one hand, the complex and heterogeneous appearance of buildings leads to internal variability. On the other hand, the mixed backgrounds and other objects with similar spectral signatures further limit the class separability.

Nowadays, convolutional neural networks (CNNs) have been widely used for remote sensing tasks \cite{zhu2017deep} \cite{shi2018building} \cite{hua2020relation} , as they surpass conventional methods in terms of accuracy of efficiency. CNNs are capable of directly learning hierarchical contextual features from the original input, which have greater generalization capabilities for the building footprint generation from remote sensing imagery. Although the existing CNNs are able to deliver very promising results \cite{shi2018building} \cite{shi2020building} \cite{li2020building} \cite{li2021building}, there remains a challenge for extracting building footprints \lia{on a} large scale. This challenge arises from that CNNs require massive annotated data to obtain strong supervisory information. However, manual annotation of reference data is a time-consuming and costly process. 

To address this issue, a straightforward idea is to utilize semi-supervised learning, which can leverage a large amount of unlabeled data and alleviate the need for labeled examples. In general, semi-supervised semantic segmentation methods are summarized into three types: weakly-supervised training-based, adversarial training-based, and consistency training-based. Nevertheless, weakly-supervised training-based methods need additional annotations, e.g. image-level labels or region-level labels. Adversarial training-based methods are able to make use of the unlabeled data but are difficult to train. Consistency training-based approaches, while not only are simple to implement, but also require no additional weakly labeled examples. The core idea of consistency training-based methods is to encourage the network to give consistent outputs for unlabeled inputs that are perturbed in various ways, thus, improving the generalization of the network \cite{athiwaratkun2018there}.

The state-of-the-art consistency training-based methods exploit the teacher-student framework \cite{qi2020small}. Specifically, a student model is applied to the unlabeled sample, while a teacher model is applied to a perturbed version of the same sample. Afterward, the consistency is imposed between the outputs of two models to improve the performance of the student model \cite{qi2020small}. However, there is still a certain gap in performance between these two models when the outputs are not completely correct during training.  Inspired by \cite{johnson2016perceptual} that feature maps can capture more discriminative contextual information, we further improve the performance of consistency training by proposing a new consistency loss that measures the discrepancy between both feature maps and outputs of student model and those of teacher model. By doing so, it can offer a strong constraint to regularize the learning of the network.

The effectiveness of consistency training-based approaches depends heavily on the behavior of the data distribution, i.e., the cluster assumption, where the classes must be separated by low-density regions. However, the low-density regions separating the classes are not within the inputs, which offers an explanation for why semi-supervised is a challenging problem for semantic segmentation \cite{french2020semi}. \cite{ouali2020semi} observes that for natural images low-density regions separating the classes are present at the encoder's output, thus, proposing to assign the perturbation at this position. However, for remote sensing imagery with low spatial resolution, we observe the presence of low-density regions separating the classes is within the intermediate feature representations in the encoder rather than the encoder's input or output. Motivated by this observation, in this work, we propose to enforce the consistency over the perturbation applied to feature representations at a certain depth within the encoder, where this depth should be in line with the spatial resolution of remote sensing imagery and the mean size of individual buildings in the study area.

Specifically, we consider a shared encoder and a main decoder that are trained together using the labeled examples. To leverage unlabeled data, we then consider an auxiliary decoder whose inputs are perturbed versions of the shared encoder's output. The consistency is imposed between outputs and feature maps of the main decoder and those of the auxiliary decoder. By doing so, the shared encoder’s representation is enhanced by using the additional training signal extracted from the unlabeled data. 

This work's contributions are threefold.

(1) We propose a semi-supervised network for building footprint generation, which has not been adequately addressed in the current literature. When the annotated samples are insufficient, the proposed method can leverage a large amount of unlabeled data to improve the performance of a model.

(2) Our proposed method integrates the consistency training of features and outputs into a unified objective function, which formulates an efficient end-to-end training framework. Compared with other competitors, our approach gains significant improvements. 

(3) Observing that the low-density regions separating the classes are within the intermediate feature representations in the encoder, we propose an instruction, in which the perturbation is applied on the feature representations at a certain depth within the encoder according to the spatial resolution of input remote sensing imagery and the mean size of individual buildings in the study area.

The remainder of the paper is organized as follows. Related work is reviewed in Section \ref{sec:rel}. Section \ref{sec:met} details the proposed network for building footprint generation. The experiments are described in Section \ref{sec:exp}. Results and Discussions are provided in Section \ref{sec:res} and \ref{sec:dis}, respectively. Eventually, Section \ref{sec:con} summarizes this work.

\section{Related Work}
\label{sec:rel}
\subsection{Building Footprint Generation}
A tremendous amount of remote sensing imagery can be collected with recent technological advances, providing huge potential for mapping buildings. A variety of methods have been proposed to generate building footprints from remote sensing imagery.

Early studies can be categorized into four types: geometrical primitive-based, index-based, segmentation-based, and classification-based methods. The geometrical primitive-based methods \cite{qin2018accurate} first extract geometric primitives (e.g., building edges and corners) and then group them to form building hypotheses. In the index-based methods \cite{huang2011multidirectional}, an index is designed to discriminate buildings from other objects. Afterward, buildings are extracted by selecting an empirical threshold. By utilizing over-segmentation algorithms, the segmentation-based methods \cite{ok2013automated} aims at partitioning an image into different segments, so-called superpixels, and identify those belonging to buildings. In the classification-based methods \cite{turker2015building},  spectral and/or spatial features of each pixel are taken as input of classifiers to differentiate building from other classes. Nonetheless, a general limitation of these methods is that they rely heavily on manually defined rules and handcrafted features, resulting in a decrease in accuracy and efficiency.

In the past few years, deep learning-based methods have shown remarkable performance on this task, as discriminative features from raw images can be automatically and adaptively learned. Early methods \cite{mnih2013machine} \cite{alshehhi2017simultaneous} employ a patch-wise classification framework, and assign the label to each pixel according to the class of its enclosing patch. However, the large overlap among patches leads to redundant operation and low efficiency. Therefore, semantic segmentation networks that can efficiently perform pixel-wise segmentation, becomes more popular in the task of building footprint generation \cite{maggiori2017dataset} \cite{liu2021building} \cite{huang2021lightweight}  \cite{li2020building} \cite{li2021building} \cite{guo2020scene} \cite{zhu2020map} \cite{jung2021boundary} \cite{lee2021boundary} \cite{peng2021full}. The commonly used network architectures involve fully convolutional networks (FCNs) \cite{long2015fully}  and encoder-decoder based architectures (e.g., DeepLabv3+\cite{chen2018encoder} Efficient-UNet \cite{baheti2020eff}, FC-DenseNet \cite{jegou2017one}). \lia{In order to take the characteristics of buildings in remote sensing imagery into account, some methods (e.g., ESFNet \cite{lin2019esfnet}, MA-FCN \cite{wei2019toward}, HA U-Net \cite{xu2021ha}, and Multi-task \cite{bischke2019multi}) have made some specific adaptations to these network architectures, e.g., attention block and multi-scale feature aggregation.} More recently, instance segmentation networks are exploited to delineate individual building instances in several novel studies \cite{liu2020multiscale} \cite{li2020instance}. Instance segmentation networks can not only assign a semantic label to each pixel with the class of its enclosing object but also distinguish different instances. The commonly used instance segmentation architecture for this task is Mask R-CNN \cite{he2017mask}.
\subsection{Semi-Supervised Semantic Segmentation}
Deep learning methods require strong supervisory information for network training, however, the collection of large volumes of annotated data is time-consuming and costly. Especially for the task of semantic segmentation, the acquisition of pixel-level labels is more expensive and laborious. Therefore, semi-supervised learning is favored in this task, and it can leverage a large amount of unlabeled data to compensate for limited supervisory information. In general, semi-supervised semantic segmentation methods are summarized into three types: weakly-supervised training-based, adversarial training-based, and consistency training-based.

\lia{Weakly-supervised training-based methods \cite{wei2017object} \cite{wei2018revisiting} \cite{lee2019ficklenet} \cite{huang2018weakly} integrate weakly-supervised learning in their approaches. Apart from the limited pixel-level labels, they still require weaker labels that can be regarded as supervisory information for network training. For the application of building footprint generation, weaker labels include image-level labels, bounding boxes, and point labels. The image-level label has two classes, where “building” refers to the images occupying building pixels more than a certain amount of the total pixels, and “non-building” corresponds to images without building pixels \cite{chen2020spmf} \cite{li2021effectiveness}. In \cite{rafique2019weakly}, bounding box annotations are utilized to generate probabilistic masks using bivariate Gaussian distribution for every image. Point labels (two points inside and outside each small building, respectively) are employed in \cite{lee2021weakly}, which is helpful to detect small buildings. Nevertheless, weakly-supervised training-based methods fail to take advantage of massive unlabeled data.} Adversarial training-based methods \cite{hung2018adversarial} \cite{souly2017semi} are able to exploit unlabeled samples, which adapt generative adversarial networks (GANs) \cite{goodfellow2014generative} for semi-supervised semantic segmentation. Both the generator and the discriminator are first trained by labeled samples. Afterward, the generator outputs the segmentation masks of unlabeled images, while the discriminator distinguishes trustworthy regions in their predicted results to provide additional supervisory signals. \lia{Considering that the adversarial training strategy may be insufficient to guide network training, pseudo labels are generated by selecting high-confident segmentation predictions for unlabeled images \cite{yao2021weakly}. Afterward, pseudo-building masks are incorporated to expand the training data and the generator is retrained. However, adversarial training-based methods are very hard to train due to the instability of GANs \cite{arjovsky2017wasserstein}.} By contrast, consistency training-based methods not only can leverage unlabeled images to improve the performance of the segmentation network but also are simple and efficient to implement. The goal of consistency training is to enforce the consistency of the model’s predictions for unlabeled inputs that are applied by small perturbations. By doing so, the robustness of the learned model will be enhanced

Recently, several consistency training-based methods are proposed for the task of semi-supervised semantic segmentation, e.g., CutMix \cite{french2020semi} and CCT \cite{ouali2020semi}. CutMix \cite{french2020semi} applies the perturbations to the raw input and uses MixUp \cite{zhang2017mixup} to enforce the consistency between the mixed outputs and the outputs from the mixed inputs. CCT \cite{ouali2020semi} imposes an invariance of the model’s outputs over small perturbations applied to the encoder’s output. \lia{In the remote sensing community, two consistency training-based methods have been proposed for the application of building footprint generation, i.e., CR \cite{wang2020semi} and PiCoCo \cite{kang2021picoco}. Color jitter and random noise are chosen as the perturbation for CR \cite{wang2020semi}, and are applied to the raw input. Then, the consistency of their outputs is enforced. PiCoCo \cite{kang2021picoco} is also an input perturbation method, which augment the input images randomly and impose the consistency constraint between the predictions of augmented images. In addition, it implements contrast learning on labeled images, which can regularize the compactness of intra- and interclass latent representation space \cite{kang2021picoco}.}

\lia{However, these consistency training-based methods still have two limitations.} On the one hand, these methods ignore the rich information encoded in feature maps and generally impose consistency only over the outputs of the models. On the other hand, they add perturbations over the raw input or encoder's output for all types of data, failing to take the characteristics of target objects into consideration when selecting the optimal position to apply perturbations.  

\begin{figure*}[!t]
 \begin{center}
  \includegraphics[width=0.8\linewidth]{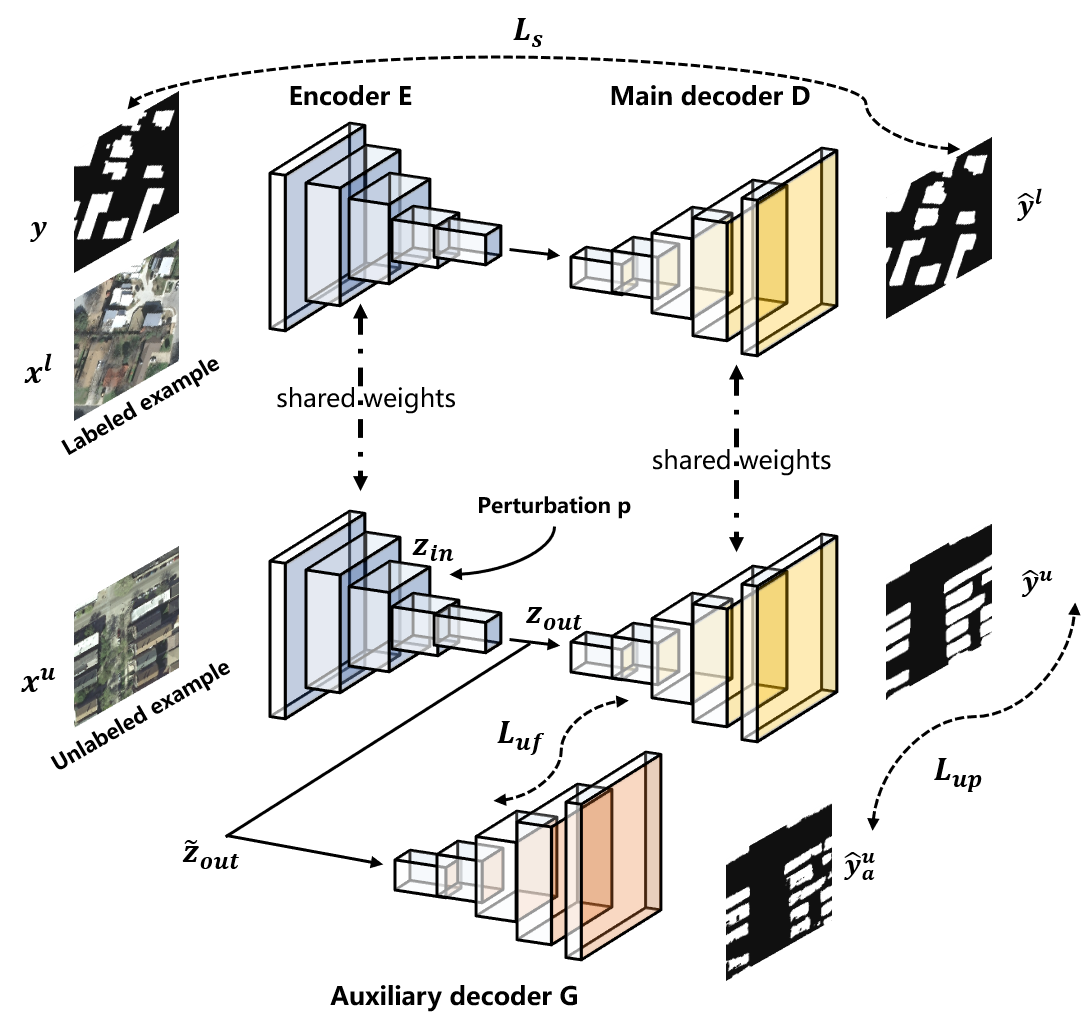}
  \caption{Overview of the proposed semi-supervised building footprint generation network.}
  \label{Fig. 2}
  \end{center}
\end{figure*}

\section{Methodology}
\label{sec:met}
In this section, consistency training-based methods are first introduced. Afterward, the proposed framework in the end-to-end network learning procedure is described. Finally, we propose an instruction to assign perturbation for the task of building footprint generation, which is based on our observation and analysis of cluster assumption.
\subsection{Consistency Training-based Methods}
Given a small set of $n$ input-target pairs $S_l=\{(x_1^l,y_1),...,(x_n^l,y_n)\}$ sampled from an unknown joint distribution $\beta (x,y)$, the goal of supervised learning is to derive a prediction function $f_\theta (x)$ parametrized  by $\theta$, and this prediction function is able to assign the correct target $y$ to an unseen sample from $\beta(x)$. In semi-supervised learning, a larger set of $m$ unlabeled examples $S_u=\{x_1^u,...,x_m^u\}$ is additionally provided. Semi-Supervised learning aims to derive a more accurate prediction function than what is obtained by only using $S_l$. For instance, additional structure about the input distribution $\beta(x)$ can be learned from $S_u$ to produce a estimate of the decision boundary, which makes a better separation of samples into different classes \cite{oliver2018realistic}.

Consistency training-based methods follow an intuitive goal to perform semi-supervised learning: when a perturbation is assigned to the data points $x\in S_u$ as $\hat{x}$, the output of $f_\theta (x)$ should not be significantly changed. Therefore, the objective of consistency training-based methods is to minimize the following loss function:

\begin{equation}
\begin{aligned}
\label{eqn:1}
       L=L_s+\lambda_u \cdot L_{cons} \,,
\end{aligned}
\end{equation}
where $L_s$ is a supervised loss on labeled data. $\lambda_u$ is a weighting function to control the importance of a consistency loss term $L_{cons}$ which is formalized as:

\begin{equation}
\begin{aligned}
      L_{cons}=\mathbf{T}(f_\theta(x),f_\theta(\hat{x})) \,,
\end{aligned}
\end{equation}
where $\mathbf{T} (.,.)$ measures a discrepancy between the outputs of the prediction functions. In this regard, the unlabeled data can be leveraged to find a smooth manifold where the dataset lies \cite{belkin2006manifold}. 

Different settings in assigning perturbation or minimizing the $L_{cons}$ lead to a wide variety of approaches for semi-supervised classification, e.g., Virtual Adversarial Training (VAT) \cite{miyato2018virtual} and Interpolation Consistency Training (ICT) \cite{verma2019interpolation}, and those from semi-supervised semantic segmentation, e.g., CutMix\cite{french2020semi}, CCT \cite{ouali2020semi}, \lia{CR \cite{wang2020semi}, and PiCoCo \cite{kang2021picoco}}. These methods are conducted in teacher-student frameworks, where a teacher model is first constructed from data perturbation, and then the output of the teacher model on unlabeled data is utilized to supervise a student model \cite{qi2020small}. However, they have not fully leveraged the information of the teacher model. This is because they fail to use intermediate feature maps of the teacher model that can also be regarded as knowledge to guide the learning of the student model. Therefore, a more precise consistency towards the underlying invariance of features and outputs between the student model and the teacher model is preferable in our research.

\subsection{Proposed Framework in End-to-End Network Learning}
Recently, the perceptual mechanism has achieved promising results for image reconstruction \cite{johnson2016perceptual}, and they make use of the extracted high-level feature maps to improve the network performance. Inspired by it, we propose to impose consistency on both features and predictions for the training of unlabeled data, which is capable of fully harnessing information in deep features and output predictions. As a consequence, our network can guarantee that the deep feature maps are consistent, alleviating the loss of detailed information during network training.

As shown in Fig. \ref{Fig. 2}, the proposed framework is composed of a shared encoder $E$, a main decoder $D$, and an auxiliary decoder $G$. The segmentation network $F$ is constituted as $F = E \circ D$ and is trained on the labeled set in a fully supervised manner. The auxiliary network $A = E \circ G$ is trained on the unlabeled examples by enforcing the consistency of both features and outputs between $D$ and $G$. $D$ takes as input the encoder’s output $\mathbf{z}_{out}$, but $G$ is fed with its perturbed version $\tilde{\mathbf{z}}_{out}$, in which the perturbation $p$ is applied to feature representations $\mathbf{z}_{in}$ at a certain depth within $E$. By doing so, the representation learning of $E$ can be further improved by unlabeled examples, and subsequently, that of the segmentation network $F$.

For each iteration of training, a labeled input image $x^l$ and its label $y$ together are sampled together with an unlabeled image $x^u$. Both $x^l$ and $x^u$ are passed through $E$ and $D$, obtaining two main predictions $\hat{y}^l$ and $\hat{y}^u$, respectively. The supervised loss $L_s$ is computed with $y$ and $\hat{y}^l$. For $x^u$, the perturbation $p$ is applied to $\mathbf{z}_{in}$ with $\mathbf{z}_{in}$ being its feature representation within $E$ and its output from $E$ is $\tilde{\mathbf{z}}_{out}$. Afterward, an auxiliary prediction $\hat{y}_a^u$ is generated from $G$ using the $\tilde{\mathbf{z}}_{out}$. The consistency loss $L_{cons}$ consists of two parts $L_{uf}$ and $L_{up}$, where $L_{uf}$ is computed between the features of $G$ and those of $D$, and $L_{up}$ is computed between the outputs of $G$ and that of $D$. 

In the proposed approach, $S_l$ and $S_u$ are jointly trained by minimizing a global loss function $L$ as Eq. \ref{eqn:1}. Following \cite{laine2016temporal}, $\lambda_u$ is set to ramp up starting from zero along a Gaussian curve up to a fixed weight $\alpha$, which can avoid the use of the initial noisy output from the main encoder. The total loss $L$ is derived and back-propagated to train the segmentation network $F$ and the auxiliary network $A$. \lib{Note that $L_{cons}$ is not backpropagated through $D$, and $D$ is trained only by labeled examples. By doing so, $D$ is only trained on original input data. This is helpful from two aspects. On the one hand, it can avoid collapsing solutions. If $L_{cons}$ is backpropagated through both main decoder D and auxiliary decoder $G$, main decoder $D$ will collapse since $L_{cons}$ will be minimized if predictions of both $D$ and $G$ are zeros. On the other hand, the method can be better adapted to the test stage since no perturbation is applied to test images.}

For the labeled set, a supervised loss $L_s$ is exploited to train the segmentation network $F$. In order to avoid overfitting, an annealed version of the bootstrapped Cross-Entropy loss \cite{ouali2020semi} is chosen to compute the supervised loss $L_s$, and it is denoted as:

\begin{equation}
\begin{aligned}
    L_s=  \frac{1}{|S_l|}\sum_{x_i^l,y_i\in S_l}\{F(x_i^l)<\eta\}\mathbf{H}(y_i, F(x_i)) \,,
\end{aligned}
\end{equation}  
where $F(x_i)$ is the output probability from $F$ for \lia{a labeled example} $x_i$, $y_i$ is its ground reference label, and $\mathbf{H}(.,.)$ is the cross entropy-based loss. \lib{In semi-supervised learning, the model is often overfitted to the limited amount of labeled data while being under-fitted to the unlabeled data. To address this issue, a labeled example is utilized only if the model’s confidence in it is lower than a predefined threshold $\eta$. In other words, $L_s$ is computed only over the pixels with a probability less than the threshold $\eta$ that serves as a ceiling to prevent over-training on easy labeled data \cite{xie2019unsupervised}.} Following \cite{ouali2020semi}, we gradually increase $\eta$ from 0.5 to 0.9 during the beginning of training.

For an unlabeled example $x_i^u$, $\mathbf{z}_{out}$ is derived as the output from the shared encoder $E$. One contribution in our approach is to apply the perturbation to the feature representation $\mathbf{z}_{in}$ for $x^u$ within the encoder $E$ according to our proposed instruction. Afterward, the perturbed feature representations $\tilde{\mathbf{z}}_{in}$ will be fed to the subsequent layers in the encoder to generate the perturbed encoder's output $\tilde{\mathbf{z}}_{out}$. Finally, $\mathbf{z}_{out}$ and $\tilde{\mathbf{z}}_{out}$ are taken as input for $D$ and $G$, respectively.

The training objective of the unlabeled set is to minimize \lia{a} consistency loss $L_{cons}$, which is defined as:
\begin{equation}
\begin{aligned}
     L_{cons}=L_{up}+\omega_u \cdot L_{uf} \,,
\end{aligned}
\end{equation}  
where $L_{uf}$ and $L_{up}$ measure the discrepancy between the features and outputs of $D$ and those of $G$, respectively. $\omega_u$ is a hyperparameter to introduce a weight to model the relative importance of two losses. More specifically, $L_{up}$ is defined as:
\begin{equation}
\begin{aligned}
     L_{up}=\frac{1}{|S_u|}\sum_{x_i^u\in S_u} \mathbf{T}(D(\mathbf{z}_{out}),G(\tilde{\mathbf{z}}_{out})) \,,
\end{aligned}
\end{equation} 
with $\mathbf{T}(.,.)$ as mean squared error-based loss. 

Note that a contribution of our approach is that a loss term $L_{uf}$ is introduced into the proposed network by imposing the consistency on features between the main decoder and auxiliary decoder, which is able to harness the detailed information in the feature maps. Let $\phi_j(q)$ be the activations of the $j$th layer of the network $\phi$ when processing the input $q$. For $D$ and $G$, $D_j(\mathbf{z}_{out})$ and $G_j(\tilde{\mathbf{z}}_{out})$ will be the corresponding feature maps at $j$th depth in the decoder. Here, $j$ represents the position where upsampling operations are applied in the decoder. Then, $L_{uf}$ is denoted as:
\begin{equation}
\begin{aligned}
     L_{uf}=\frac{1}{|S_u|}\sum_{x_i^u\in S_u} \sum_{j=1}^{J}\mathbf{T}(D_j(\mathbf{z}_{out}),G_j(\tilde{\mathbf{z}}_{out})) \,,
\end{aligned}
\end{equation} 
where $J$ is the total number of depth in the decoder. In other words, $J$ represents how many upsampling operations are applied in the decoder.
The proposed semi-supervised method can be summarized by the following Algorithm \ref{al1}:

\begin{algorithm}
 \caption{Algorithm for Feature and Output Consistency Training}
 \begin{algorithmic}[1]
 \renewcommand{\algorithmicrequire}{\textbf{Input:}}
 \renewcommand{\algorithmicensure}{\textbf{Require:}}
 \REQUIRE Labeled image $x^l$ and pixel-level label $y$, as well as unlabeled image $x^u$
 \ENSURE  Shared encoder $E$, main decoder $D$ with the total depth number $J$, and auxiliary decoder $G$
  \STATE Forward $x^l$ through $E$ and $D$: $\hat{y}^l=D(E(x^l))$
  \STATE Forward $x^u$ through $E$: $\mathbf{z}_{out}=E(x^u)$
   \STATE Generate the main decoder's feature maps for $\mathbf{z}_{out}$: 
   \\ \FOR {$j = 1$ to $J$} 
    \STATE Derive $D_j(\mathbf{z}_{out})$
       \ENDFOR
    \STATE Generate the main decoder's output for $\mathbf{z}_{out}$: \\Derive $D(\mathbf{z}_{out})$  
    \STATE Forward $x^u$ through $E$ and apply a noise perturbation $\mathbf{N}$ to feature representations $\mathbf{z}_{in}$: 
     $\tilde{\mathbf{z}}_{in}=(\mathbf{z}_{in}\odot \mathbf{N})+\mathbf{z}_{in}$
    \STATE Forward $\tilde{\mathbf{z}}_{in}$ through the subsequent layers in $E$ to generate the perturbed encoder’s output $\tilde{\mathbf{z}}_{out}$
   \STATE Generate the auxiliary decoder's feature maps for $\tilde{\mathbf{z}}_{out}$: 
   \\ \FOR {$j = 1$ to $J$} 
    \STATE Derive $G_j(\tilde{\mathbf{z}}_{out})$
   \ENDFOR
    \STATE Generate the auxiliary decoder's output for $\tilde{\mathbf{z}}_{out}$: \\Derive $G(\tilde{\mathbf{z}}_{out})$  
    \STATE Training the network.
    \\$L_s=\{\hat{y}^l<\eta\}\mathbf{H}(y, \hat{y}^l)$
    \\$L_{up}=\mathbf{T}(D(\mathbf{z}_{out}),G(\tilde{\mathbf{z}}_{out}))$
    \\ $L_{uf}= \sum_{j=1}^{J}\mathbf{T}(D_j(\mathbf{z}_{out}),G_j(\tilde{\mathbf{z}}_{out}))$
    \\Update network by $L=L_s+\lambda_u \cdot (L_{up}+\omega_u \cdot L_{uf})$
 \end{algorithmic} 
 \label{al1}
 \end{algorithm}
 
\subsection{An Instruction to Assign Perturbation for the Task of Building Footprint Generation}

The effectiveness of consistency training-based methods relies on the cluster assumption, i.e., two samples belonging to the same cluster in the input distribution are likely to have the same label \cite{chapelle2005semi}. In this case, the decision boundary should lie in the low-density regions \cite{luo2018smooth}. In other words, if a decision boundary crosses a high-density region, it will divide a cluster into two different classes, which violates the cluster assumption. From the formal analysis, the expected value of $L_{cons}$ is proportional to the squared magnitude of the Jacobian of the network’s outputs with respect to its inputs \cite{athiwaratkun2018there}. Therefore, minimizing $L_{cons}$ indicates that the decision function in the regions of unsupervised samples will be flattened, and the decision boundary will be moved into the vicinity of low sample density \cite{french2020semi}.

\begin{figure}[!t]
 \begin{center}
  \includegraphics[width=1.0\linewidth]{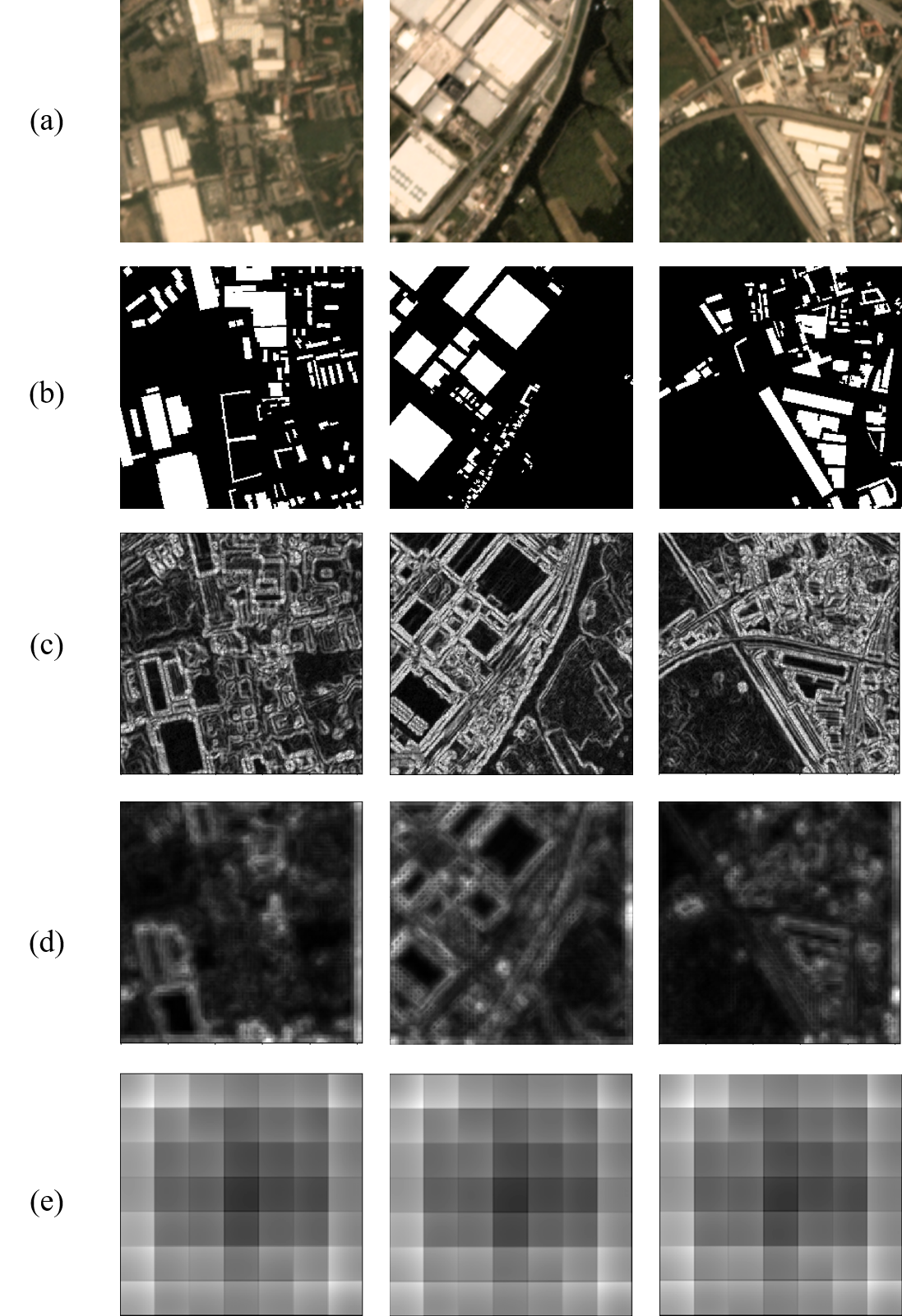}
  \caption{\lia{The cluster assumption in consistency training-based methods for building footprint generation. Examples from (a) Planet satellite imagery (3m/pixel), (b) pixel-level labels, as well as local variations at (c) encoder's input, (d) intermediate layer in the encoder, and (e) encoder's output. Bright regions indicate large variation.}}
  \label{Fig. pd}
  \end{center}
\end{figure}

The cluster assumption has inspired many recent consistency training-based methods for semi-supervised semantic segmentation \cite{french2020semi} \cite{ouali2020semi} which propose to assign the perturbation to the raw input or encoder's output. However, they are not suitable for the task of building footprint generation, as the characteristics of both building objects and remote sensing imagery haven't been taken into account. Therefore, we propose an instruction to assign perturbation for this task, which is inspired by the observation and analysis of the cluster assumption in building footprint generation from remote sensing imagery. In order to examine the cluster assumption, the local variations at an encoder depth $d$ are measured between the value of each pixel and its local neighbors, and local variations with high values depict the presence of low-density regions \cite{french2020semi}. Here, $d$ represents the position where how many downsampling operations are applied in the encoder. For instance, when $d=1$, the spatial size (i.e., height and width) of feature representation is half of that of the raw input. Similarity, when $d=2$, the spatial size (i.e., height and width) of feature representation is 1/4 of that of the raw input. Following \cite{ouali2020semi}, the average Euclidean distance at each spatial location and its 8 intermediate neighbors is computed for the encoder's input ($d=0$), and the feature representations of both intermediate layer ($d=2$) and encoder's output ($d=5$). Both feature representations are first resampled to the input size, and then the average distance between the neighboring activations is calculated. Fig. \ref{Fig. pd} illustrates the example results for Planet satellite imagery (3 m/pixel). The feature representations from intermediate layer and encoder's output are 24-dimensional and 1280-dimensional feature vectors learned from Efficient-UNet \cite{baheti2020eff}, respectively. It can be observed that the low-density regions are not aligned with the class boundaries at the encoder's input or encoder's output, where the cluster assumption is violated. By contrast, the cluster assumption is maintained at the intermediate layer, given that the class boundaries with high average distance coincide with low-density regions. This observation may be related to the receptive field of the network. The receptive field will be enlarged when the depth increases within the encoder, but when the receptive field exceeds a certain value that is much beyond the size of target objects, it might introduce more noise for network learning \cite{luo2017understanding}. Furthermore, for remote sensing imagery with varying resolutions, the receptive fields of the network are various at the same depth within the encoder\lia{,} when the unit is meter. 

Based on the above observation and analysis, we propose an instruction to assign the perturbation. The perturbation should be added to the feature presentations at depth $d$ within the encoder according to the spatial resolution of remote sensing imagery and the mean size of individual buildings in the study area. More specifically, $d$ is computed as:

\begin{equation}
\begin{aligned}
    d= \lfloor log_2(\frac{l_{min}+l_{max}}{2r}) \rfloor \,,
\end{aligned}
\end{equation}  
where $r$ is the spatial resolution of the remote sensing imagery, $l_{min}$ and $l_{max}$ are mean values of max and min length that are derived from the ground reference of individual buildings in the study area. $\lfloor\ \rfloor$ is the rounding down function, which aims to get the largest integer that does not exceed the original value.

A noise tensor $\mathbf{N} \sim \mu (-0.3,0.3)$ of the same size as the feature presentations $\mathbf{z}_{in}$ is uniformly sampled as the perturbation $p$. It is first multiplied with $\mathbf{z}_{in}$ to adjust its amplitude, and then injected into $\mathbf{z}_{in}$ to get perturbed feature maps $\tilde{\mathbf{z}}_{in}$:
\begin{equation}
\begin{aligned}
\tilde{\mathbf{z}}_{in}=(\mathbf{z}_{in}\odot \mathbf{N})+\mathbf{z}_{in} \,,
\end{aligned}
\end{equation} 
where $\odot$ denotes element-wise multiplication. Afterward, it will be fed to the subsequent layers in the encoder to generate the perturbed intermediate representation $\tilde{\mathbf{z}}_{out}$ of the unlabeled input sample $x^u$.

\section{Experiment}
\label{sec:exp}
\subsection{Dataset}

The effectiveness of the proposed method is validated on three datasets with different spatial resolutions, i.e., \lia{Planet dataset \cite{planet}, Massachusetts dataset \cite{mnih2013machine}, and Inria dataset \cite{maggiori2017dataset}}.

\begin{figure}[!t]
 \begin{center}
  \includegraphics[width=1.0\linewidth]{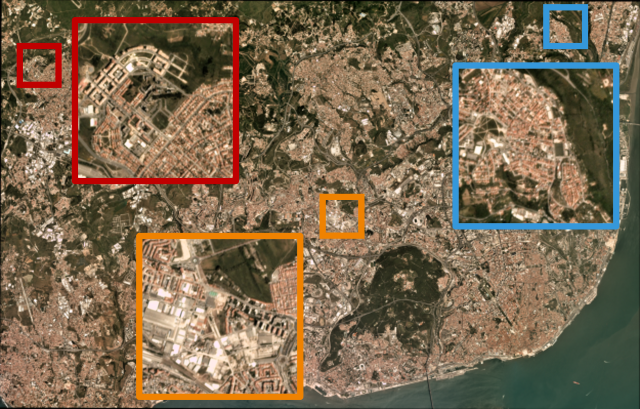}
  \caption{\lia{The satellite imagery of Lisbon in the Planet dataset (spatial resolution: 3m/pixel) and three zoomed in areas.}}
  \label{Fig. 3}
  \end{center}
\end{figure}

1) Planet dataset: In this research, PlanetScope satellite imagery is collected from 8 European cities (Amsterdam, Berlin, Lisbon, Madrid, London, Paris, Milan, and Zurich) to create a Planet dataset. The PlanetScope satellite images have three bands (i.e., red, green, blue) at a spatial resolution of 3 m/pixel. The corresponding building footprints that are stored as vector files are acquired from OpenStreetMap. Fig. \ref{Fig. 3} presents example imagery of Lisbon. 

\begin{figure}[!t]
 \begin{center}
  \includegraphics[width=1.0\linewidth]{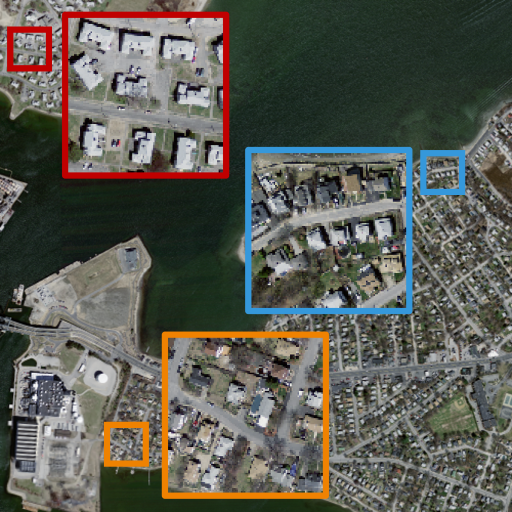}
  \caption{\lia{An aerial image in the Massachusetts dataset (spatial resolution: 1 m/pixel)and three zoomed in areas.}}
  \label{Fig. 4}
  \end{center}
\end{figure}

2) Massachusetts dataset: The Massachusetts dataset is composed of 151 tiles of aerial imagery over the city of Boston. Each aerial imagery has three bands (i.e., red, green, blue) at a spatial resolution of 1 m/pixel, and its size is $1500 \times 1500$ pixels. A sample aerial image is illustrated in Fig.  \ref{Fig. 4}. The corresponding ground reference building masks are also included in this benchmark dataset.

\begin{figure}[!t]
 \begin{center}
  \includegraphics[width=1.0\linewidth]{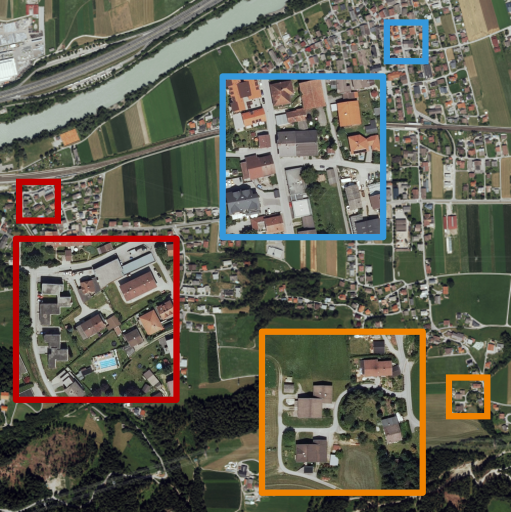}
  \caption{\lia{An aerial image in the Inria dataset (spatial resolution: 0.3 m/pixel) and three zoomed in areas.}}
  \label{Fig. 5}
  \end{center}
\end{figure}

3) Inria dataset: The Inria dataset is a benchmark dataset consisting of 360 large-scale aerial images, in which each image is of the size of $5000 \times 5000$ and has three bands (i.e., red, green, blue) at a spatial resolution of 0.3 m/pixel. A sample aerial image is showed in Fig.  \ref{Fig. 5}. The ground reference building masks of this dataset are only publicly released for five cities (Austin, Chicago, Kitsap County, Western Tyrol, and Vienna).

\begin{table}[!t]
\centering
\renewcommand{\arraystretch}{1.0}
 \caption{The statistics of the selected datasets utilized in this research.}
\begin{threeparttable}{!}{
\begin{tabular}{P{1.5cm}|P{2.4cm}|P{0.8cm}|P{1cm}|P{0.8cm}}
\Xhline{3\arrayrulewidth}

   \hline\hline
      Dataset & City & \multicolumn{3}{c}{The number of patches} \\
      \cline{3-5}
      ~ & ~  & train & validation & test\\
      \hline
      \multirow{8}{*}{Planet dataset} & Amsterdam & \multirow{8}{*}{4800} & \multirow{8}{*}{1600} & \multirow{8}{*}{2400}\\
      \cline{2-2}
      ~ & Berlin & ~ & ~ & ~\\
      \cline{2-2}      
      ~ & Lisbon & ~ & ~ & ~\\
      \cline{2-2}
      ~ & Madrid & ~ & ~ & ~\\
      \cline{2-2}
      ~ & London & ~ & ~ & ~\\
      \cline{2-2}
      ~ & Paris & ~ & ~ & ~\\
      \cline{2-2}
      ~ & Milan & ~ & ~ & ~\\
      \cline{2-2}
      ~ & Zurich & ~ & ~ & ~\\      
      \hline
      Massachusetts dataset & Boston & 3424 & 100 & 250\\
      \hline
       \multirow{5}{*}{Inria dataset} & Austin & \multirow{5}{*}{39852} & \multirow{5}{*}{6044} & \multirow{5}{*}{6044}\\
      \cline{2-2}
      ~ & Chicago & ~ & ~ & ~\\
      \cline{2-2}      
      ~ & Kitsap County & ~ & ~ & ~\\
      \cline{2-2}
      ~ & Western Tyrol & ~ & ~ & ~\\
      \cline{2-2}
      ~ & Vienna & ~ & ~ & ~\\
\Xhline{3\arrayrulewidth}
\end{tabular}}
\end{threeparttable}
 \label{Tab.s0}
 \end{table}

\begin{table*}[!t]
\centering
\renewcommand{\arraystretch}{1.0}
 \caption{\lia{The settings of all methods utilized in this research. $\lambda_u$ and $\omega_u$ represent the weights of consistency loss term and feature consistency loss term, respectively.}}
\begin{threeparttable}{!}{
\begin{tabular}{P{7cm}|P{0.8cm}|P{0.8cm}|P{7cm}}
\Xhline{3\arrayrulewidth}

   \hline\hline
       Method & $\lambda_u$ & $\omega_u$ & The position of the assigned perturbation \\
      \hline
      Supervised Learning (SL) & $=0$  & $=0$ & -\\
      \hline
      Supervised Learning + Data Augmentation (SL+DA) & $=0$  & $=0$ & -\\  
      \hline
      ICT \cite{verma2019interpolation} & $>0$  & $=0$ & encoder's input \\
      \hline
      VAT \cite{miyato2018virtual} & $>0$  & $=0$ & encoder's input \\
      \hline
      CutMix\cite{french2020semi} & $>0$  & $=0$ & encoder's input \\
      \hline
      CCT \cite{ouali2020semi} & $>0$  & $=0$ & encoder's output \\
      \hline
      \lia{CR \cite{wang2020semi}} & \lia{$>0$} & \lia{$=0$} & \lia{encoder's input} \\
      \hline
      \lia{PiCoCo \cite{kang2021picoco}} & \lia{$>0$}  & \lia{$=0$} & \lia{encoder's input} \\
      \hline
      Propose method  & $>0$  & $>0$ & encoder's intermediate feature representations \\
\Xhline{3\arrayrulewidth}
\end{tabular}}
\end{threeparttable}
 \label{Tab.sm}
 \end{table*}

For all three datasets, all remote sensing images and ground-truth building masks are cut into small patches with the size of $ 256 \times 256$ pixels. For the Planet dataset, we have manually selected 1100 pairs of proper patches for each of eight European cities. The selected pairs are then separated into three parts, and the ratio of train, validation, and test set is 6:2:3. Data split in the Inria dataset is according to the setup in \cite{maggiori2017dataset} \cite{bischke2019multi}. More specifically, for each city, images with ids 1-5 are used for validation, and the remaining 31 images are for training. The statistics are derived from the validation set. The training/validation/test split of the Massachusetts dataset follows  \cite{mnih2013machine}, where 137 tiles are used for training, 4 tiles are for validation, and the remaining 10 tiles are used to test models. The numbers of patches collected from each dataset for network training, validation, and test are reported in Table \ref{Tab.s0}.

\subsection{Experiment Setup}
\lia{Since the semantic segmentation network is an essential part of our approach, we first investigate which CNN model (i.e., Efficient-UNet \cite{baheti2020eff}, FC-DenseNet \cite{jegou2017one}, DeepLabv3+ \cite{chen2018encoder}, ESFNet \cite{lin2019esfnet}, MA-FCN \cite{wei2019toward}, HA U-Net \cite{xu2021ha}, and Multi-task \cite{bischke2019multi}) has better performance for the task of building footprint generation. The CNN model achieving the best results under the fully supervised setting is selected as the backbone.} Afterward, for each dataset, we randomly split the training data into two parts, which are labeled set and unlabeled set, and the pixel-level annotations are excluded in the unlabeled set. Under the semi-supervised setting, the ratios of labeled data to unlabeled data are set as three different ratios (e.g., 1:2, 1:5, 1:10). To validate the superiority of the proposed method, we make a comparison with other competitors, including Supervised Learning (SL), Supervised Learning + Data Augmentation (SL+DA), ICT \cite{verma2019interpolation}, VAT \cite{miyato2018virtual}, CutMix \cite{french2020semi}, CCT \cite{ouali2020semi}, \lia{CR \cite{wang2020semi} and PiCoCo \cite{kang2021picoco}.} The settings of $\lambda_u$, $\omega_u$ being the weights of consistency loss term and feature consistency loss term, and the position of the assigned perturbation in different methods are shown in Table \ref{Tab.sm} for a better understanding of their differences. Furthermore, the effectiveness of our proposed feature and output consistency, being imposed between the main decoder and the auxiliary decoder, is analyzed. The position within the encoder to apply perturbation is also carefully investigated for different datasets. \lia{Finally, we explore whether the auxiliary decoder is able to improve the performance of the proposed method.}

\subsection{Training Details}
Our experiments are conducted within a Pytorch framework on an NVIDIA Tesla with 16 GB of memory. For all methods, the optimizer is stochastic gradient descent (SGD) with a learning rate of 0.1 and a momentum of 0.9, and the training batch size is set as 4. Detailed configurations of all methods included in our experiments are listed as follows:

(1) Efficient-UNet \lia{\cite{baheti2020eff}}: EfficientNet\cite{tan2019efficientnet} is adopted as the encoder to learn feature maps. The decoder is comprised of five transposed convolutional layers that upsample the convolved image to predict segmentation masks.

(2) DeepLabv3+ \lia{\cite{chen2018encoder}}: The feature extractor in DeepLabv3+ is the Xception model \cite{chollet2017xception}.

(3) FC-Densenet \lia{\cite{jegou2017one}}: Both the encoder and decoder in FC-DenseNet are composed of five dense blocks, and each dense block has five convolutional layers.

\lia{(4) ESFNet \cite{lin2019esfnet}: This method employs Separable Factorized Residual Block (SFRB) as the core module. The encoder is composed of 16 blocks, where 3 blocks are downsampling blocks and 13 blocks are SFRB. The decoder consists of 7 blocks for transposed convolutions and SFRB.}

\lia{(5) MA-FCN \cite{wei2019toward}: This approach has proposed a feature fusion structure to aggregate multi-scale feature maps. It utilizes a Feature Pyramid Network (FPN) \cite{lin2017feature} -based structure as the backbone where the encoder is a four-layer VGG-16 \cite{simonyan2014very} architecture and a corresponding decoder implements lateral connections between them.}

\lia{(6) HA U-Net \cite{xu2021ha}: The encoder of this network adopts ResNet34 \cite{he2016deep}. The decoder is comprised of four modules that include up-sampling module, attention module, overall nesting module, and auxiliary loss module.}

\lia{(7) Multi-task \cite{bischke2019multi}: This method is based on SegNet \cite{badrinarayanan2017segnet}. It first adds one convolutional layer after the decoder to learn the distance to the border of buildings. Afterward, this learned distance mask and feature maps produced by the decoder are concatenated and fed into another convolutional layer to learn the final building masks.}

(8) Proposed method: The hyperparameter $\alpha$ in the unsupervised loss weighting function $\lambda_u$ is set as 0.6. The loss term weighting parameter of feature consistency $\omega_u$ is chosen as 0.2. The network architectures of $F$ and $A$ are the same as that of the backbone.

(9) SL: The backbone is learned from labeled samples. Note that unlabeled samples are not considered during training.

(10) SL+DA: Following \cite{huang2016building}, data augmentation is first performed by randomly horizontally or vertically flipping, or rotating the image patches before training. Afterward, the backbone is trained on labeled samples.

(11) ICT \cite{verma2019interpolation} and VAT \cite{miyato2018virtual}: Following \cite{french2020semi}, we adapt these two semi-supervised classification methods for the task of semantic segmentation. The CNN model is the same as the backbone in our proposed method.

(12) CutMix \cite{french2020semi}, CCT \cite{ouali2020semi}, \lia{CR \cite{wang2020semi} and PiCoCo \cite{kang2021picoco}}: For a fair comparison, we replace the CNN model with the same backbone in our proposed method.

\begin{table*}[!t]
\centering
\renewcommand{\arraystretch}{1.0}
 \caption{\lia{Accuracies of different semantic segmentation networks for supervised learning on Three datasets. (\%)}}
\begin{threeparttable}\resizebox{\textwidth}{!}{
\begin{tabular}{P{5cm}|P{1.9cm}|P{1.9cm}|P{1.9cm}|P{1.9cm}|P{1.9cm}|P{1.9cm}}
\Xhline{3\arrayrulewidth}

   \hline\hline
      \multirow{3}{*}{Method} & \multicolumn{2}{c|}{Planet dataset (3 m/pixel)} & \multicolumn{2}{c|}{Massachusetts dataset (1 m/pixel)} & \multicolumn{2}{c}{INRIA dataset (0.3 m/pixel)} \\

      ~ & \multicolumn{2}{c|}{4800 labeled} & \multicolumn{2}{c|}{3424 labeled} & \multicolumn{2}{c}{39852 labeled} \\
      \cline{2-7}
      ~ & F1 score & IoU & F1 score & IoU & F1 score & IoU \\
      \hline
      \begin{bfseries}Efficient-UNet \lia{\cite{baheti2020eff}}\end{bfseries} &\begin{bfseries} 59.03 \end{bfseries}& \begin{bfseries} 41.87 \end{bfseries} &\begin{bfseries} 68.70 \end{bfseries}&\begin{bfseries} 52.32 \end{bfseries} &\begin{bfseries} 85.34 \end{bfseries} &\begin{bfseries} 74.42 \end{bfseries}\\
      \hline
      DeepLabv3+ \lia{\cite{chen2018encoder}} & 45.99 & 29.86 & 65.96 & 49.21 & 80.67 & 67.61\\
      \hline
      FC-DenseNet \lia{\cite{jegou2017one}} & 55.78 & 38.68 & 68.17 & 51.87 & 84.66 & 73.41\\
      \hline
      \lia{ESFNet \cite{lin2019esfnet}} & \lia{56.55} & \lia{39.42} & \lia{67.37} & \lia{50.80} & \lia{83.65} & \lia{71.90}\\
      \hline
      \lia{MA-FCN \cite{wei2019toward}} & \lia{58.40} & \lia{41.35} & \lia{68.95} & \lia{52.62} & \lia{85.03} & \lia{74.27}\\
      \hline
      \lia{HA U-Net \cite{xu2021ha}} & \lia{53.70} & \lia{36.70} & \lia{64.87} & \lia{48.00} & \lia{84.28} & \lia{72.82}\\
      \hline
      \lia{Multi-task \cite{bischke2019multi}} & \lia{48.05} & \lia{31.62} & \lia{65.43} & \lia{48.63} & \lia{84.56} & \lia{73.26}\\
      
      \hline
\Xhline{3\arrayrulewidth}
\end{tabular}}
\end{threeparttable}
 \label{Tab.s1}
 \end{table*}

\subsection{Evaluation Metrics}
The performance of models is evaluated by two metrics: F1 score and intersection over union (IoU). They can be computed as follows.
\begin{equation}
\begin{aligned}
\textrm{F1 score}=\frac{2\times \textrm{precision}\times \textrm{recall}}{\textrm{precision}+\textrm{recall}} \,,
\end{aligned}
\end{equation}
\begin{equation}
\begin{aligned}
\textrm{IoU}=\frac{TP}{TP+FP+FN} \,,
\end{aligned}
\end{equation}
\begin{equation}
\begin{aligned}
\textrm{precision}=\frac{TP}{TP+FP} \,,
\end{aligned}
\end{equation}
\begin{equation}
\begin{aligned}
\textrm{recall}=\frac{TP}{TP+FN} \,,
\end{aligned}
\end{equation}
where $TP$ indicates the number of true positives, $FN$ is the number of false negatives, and $FP$ is the number of false positives. F1 score realizes a harmonic mean between precision and recall.

\section{Results}
\label{sec:res}

 \begin{figure*}[!t]
 \captionsetup[subfigure]{labelformat=empty}
 \begin{center}
  \subfloat[(a)]{\includegraphics[width=0.18\textwidth]{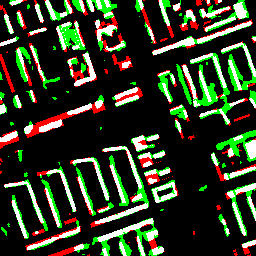}}
\hfil
   \subfloat[(b)]{\includegraphics[width=0.18\textwidth]{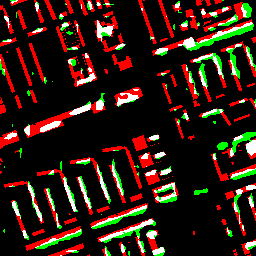}}
\hfil
   \subfloat[(c)]{\includegraphics[width=0.18\textwidth]{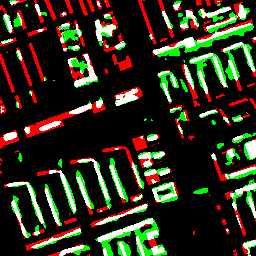}}
\hfil
  \subfloat[(d)]{\includegraphics[width=0.18\textwidth]{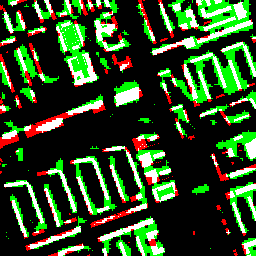}}
\vfil
   \subfloat[(e)]{\includegraphics[width=0.18\textwidth]{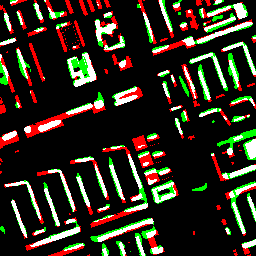}}
\hfil
   \subfloat[(f)]{\includegraphics[width=0.18\textwidth]{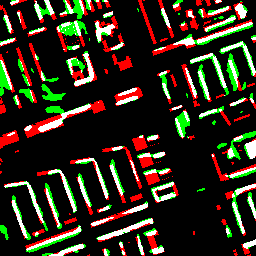}}
\hfil
  \subfloat[(g)]{\includegraphics[width=0.18\textwidth]{pic/fullsup/plahaunet.png}}
\hfil
    \subfloat[(h)]{\includegraphics[width=0.18\textwidth]{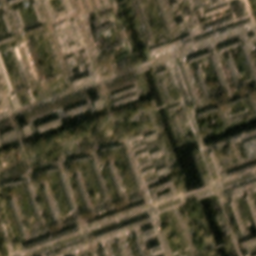}}

  \caption{\lia{Results obtained from (a) Efficient-UNet \cite{baheti2020eff}, (b) DeepLabv3+ \cite{chen2018encoder}, (c) FC-DenseNet \cite{jegou2017one}, (d) ESFNet \cite{lin2019esfnet}, (e) MA-FCN \cite{wei2019toward}, (f) HA U-Net \cite{xu2021ha}, and (g) Multi-task \cite{bischke2019multi}. (h) is satellite imagery from the Planet dataset (spatial resolution: 3 m/pixel).  Pixel-based true positives, false positives, and false negatives are marked in white, green, and red, respectively.}}
  \label{Fig. 6a}
  \end{center}
\end{figure*}

 \begin{figure*}[!t]
 \captionsetup[subfigure]{labelformat=empty}
 \begin{center}
  \subfloat[(a)]{\includegraphics[width=0.18\textwidth]{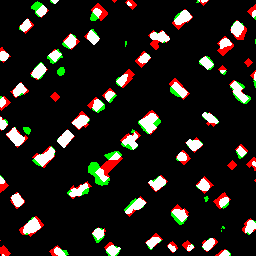}}
\hfil
   \subfloat[(b)]{\includegraphics[width=0.18\textwidth]{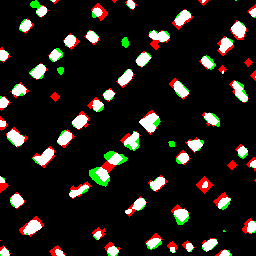}}
\hfil
   \subfloat[(c)]{\includegraphics[width=0.18\textwidth]{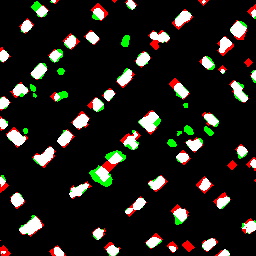}}
\hfil
  \subfloat[(d)]{\includegraphics[width=0.18\textwidth]{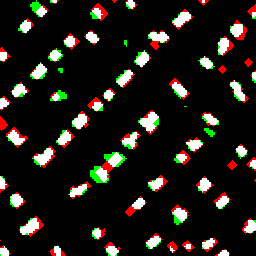}}
\vfil
   \subfloat[(e)]{\includegraphics[width=0.18\textwidth]{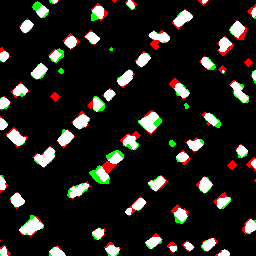}}
\hfil
   \subfloat[(f)]{\includegraphics[width=0.18\textwidth]{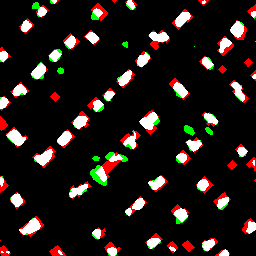}}
\hfil
  \subfloat[(g)]{\includegraphics[width=0.18\textwidth]{pic/fullsup/mashaunet.png}}
\hfil
    \subfloat[(h)]{\includegraphics[width=0.18\textwidth]{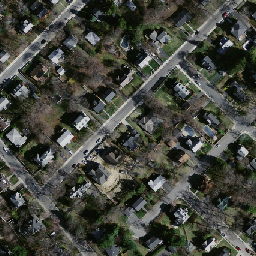}}

  \caption{\lia{Results obtained from (a) Efficient-UNet \cite{baheti2020eff}, (b) DeepLabv3+ \cite{chen2018encoder}, (c) FC-DenseNet \cite{jegou2017one}, (d) ESFNet \cite{lin2019esfnet}, (e) MA-FCN \cite{wei2019toward}, (f) HA U-Net \cite{xu2021ha}, and (g) Multi-task \cite{bischke2019multi}. (h) is satellite imagery from the Massachusetts dataset (spatial resolution: 1 m/pixel).  Pixel-based true positives, false positives, and false negatives are marked in white, green, and red, respectively.}}
  \label{Fig. 6b}
  \end{center}
\end{figure*}

 \begin{figure*}[!t]
 \captionsetup[subfigure]{labelformat=empty}
 \begin{center}
  \subfloat[(a)]{\includegraphics[width=0.18\textwidth]{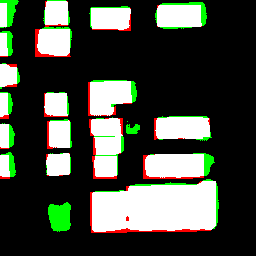}}
\hfil
   \subfloat[(b)]{\includegraphics[width=0.18\textwidth]{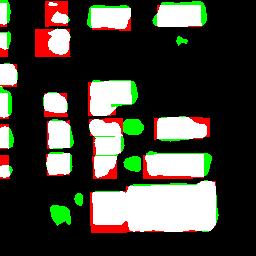}}
\hfil
   \subfloat[(c)]{\includegraphics[width=0.18\textwidth]{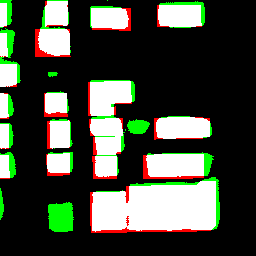}}
\hfil
  \subfloat[(d)]{\includegraphics[width=0.18\textwidth]{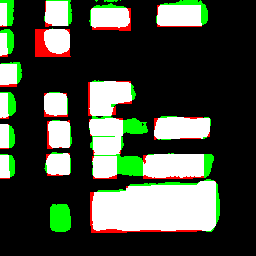}}
\vfil
   \subfloat[(e)]{\includegraphics[width=0.18\textwidth]{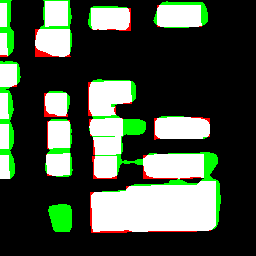}}
\hfil
   \subfloat[(f)]{\includegraphics[width=0.18\textwidth]{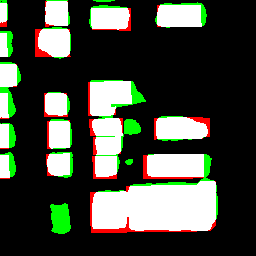}}
\hfil
  \subfloat[(g)]{\includegraphics[width=0.18\textwidth]{pic/fullsup/inrhaunet.png}}
\hfil
    \subfloat[(h)]{\includegraphics[width=0.18\textwidth]{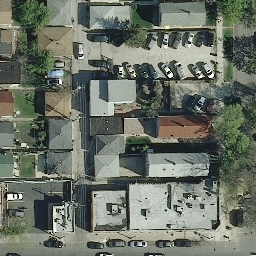}}

  \caption{\lia{Results obtained from (a) Efficient-UNet \cite{baheti2020eff}, (b) DeepLabv3+ \cite{chen2018encoder}, (c) FC-DenseNet \cite{jegou2017one}, (d) ESFNet \cite{lin2019esfnet}, (e) MA-FCN \cite{wei2019toward}, (f) HA U-Net \cite{xu2021ha}, and (g) Multi-task \cite{bischke2019multi}. (h) is satellite imagery from the Inria dataset (spatial resolution: 0.3 m/pixel).  Pixel-based true positives, false positives, and false negatives are marked in white, green, and red, respectively.}}
  \label{Fig. 6c}
  \end{center}
\end{figure*}

\subsection{Results of Different Semantic Segmentation Networks for Supervised Learning}

\lia{The comparisons among different semantic segmentation networks for supervised learning are presented in this section. Their respective performance is evaluated according to both quantitative (cf. Table \ref{Tab.s1}) and qualitative results (cf. Fig. \ref{Fig. 6a}, \ref{Fig. 6b}, and \ref{Fig. 6c}) on three datasets, respectively.} The goal of this comparison is to select the best semantic segmentation network as the backbone for different learning methods in further experiments. In this case, we can avoid potential impacts due to convolutional layers and architectural differences.

\lia{Among these semantic segmentation networks, Efficient-UNet \cite{baheti2020eff} performs better than DeepLabv3+ \cite{chen2018encoder}, FC-DenseNet \cite{jegou2017one}, ESFNet \cite{lin2019esfnet}, HA U-Net \cite{xu2021ha}, and Multi-task \cite{bischke2019multi} on all three datasets. Especially for the Planet dataset that has a relatively low spatial resolution, Efficient-UNet \cite{baheti2020eff} obtains increments of 13.04\% and 12.01\% in F1 score and IoU when compared with DeepLabv3+ \cite{chen2018encoder}. Although MA-FCN \cite{wei2019toward} is superior to Efficient-UNet \cite{baheti2020eff} on the Massachusetts dataset, Efficient-UNet surpasses it by about 0.5\% in IoU on both Planet and Inria datasets. Fig. \ref{Fig. 6c} presents a visual comparison among different methods on three datasets. For the Inria dataset with relatively high spatial resolution, some non-building objects are wrongly identified as buildings by other methods. On the contrary, Efficient-UNet \cite{baheti2020eff} is able to avoid such false alarms. The superiority of Efficient-UNet \cite{baheti2020eff} on different resolution data can be attributed to its capability of systematically improving performance with all compound coefficients of the architecture (width, depth, and image resolution) balanced \cite{baheti2020eff}. Thus, we take Efficient-UNet \cite{baheti2020eff} as the backbone in both supervised learning and semi-supervised learning approaches for further comparisons.}

\subsection{Comparison with Other Competitors}

\begin{table*}[!t]
\centering
\renewcommand{\arraystretch}{1.0}
 \caption{\lia{Accuracies of different methods on Planet dataset (3 m/pixel). (\%)}}
\begin{threeparttable}\resizebox{\textwidth}{!}{
\begin{tabular}{P{5cm}|P{1.9cm}|P{1.9cm}|P{1.9cm}|P{1.9cm}|P{1.9cm}|P{1.9cm}}
\Xhline{3\arrayrulewidth}

   \hline\hline
      \multirow{3}{*}{Method} & \multicolumn{2}{c|}{labeled:unlabeled $\approx$ 1:2} & \multicolumn{2}{c|}{labeled:unlabeled $\approx$ 1:5} & \multicolumn{2}{c}{labeled:unlabeled $\approx$ 1:10} \\
      ~ & \multicolumn{2}{c|}{(1600 labeled, 3200 unlabeled)} & \multicolumn{2}{c|}{(800 labeled, 4000 unlabeled)} & \multicolumn{2}{c}{(400 labeled, 4400 unlabeled)} \\
      \cline{2-7}
      ~ & F1 score & IoU & F1 score & IoU & F1 score & IoU \\
      \hline
      \begin{bfseries}Proposed method\end{bfseries} &\begin{bfseries} 59.35 \end{bfseries}& \begin{bfseries} 42.20 \end{bfseries} &\begin{bfseries} 56.19 \end{bfseries}&\begin{bfseries} 39.07 \end{bfseries} &\begin{bfseries} 53.78 \end{bfseries} &\begin{bfseries} 36.78 \end{bfseries}\\
      \hline
      SL & 53.15 & 36.19 & 51.80 & 34.95 & 48.03 & 31.60\\
      \hline
      SL + DA & 53.84 & 36.83 & 52.87 & 35.93 & 48.53 & 32.04\\
      \hline
      ICT \cite{verma2019interpolation} & 54.23 & 37.20 & 52.20 & 35.32 & 49.87 & 33.22\\
      \hline
      VAT \cite{miyato2018virtual} & 36.25 & 22.13 & 34.25 & 20.67 & 33.77 & 20.32\\
      \hline
      CutMix \cite{french2020semi} & 54.10 & 37.08 & 52.43 & 35.53 & 49.86 & 33.21\\
      \hline
      CCT \cite{ouali2020semi} & 56.09 & 38.97 & 53.10 & 36.15 & 50.81 & 34.06\\
      \hline
      \lia{CR \cite{wang2020semi}} & \lia{47.17} & \lia{30.86} & \lia{44.60} & \lia{28.77} & \lia{41.38} & \lia{26.08}\\  
      \hline
      \lia{PiCoCo \cite{kang2021picoco}} & \lia{54.12} & \lia{37.10} & \lia{52.43} & \lia{35.54} & \lia{46.94} & \lia{30.67}\\          
\Xhline{3\arrayrulewidth}
\end{tabular}}
\end{threeparttable}
 \label{Tab.s2}
 \end{table*}

\begin{table*}[!t]
\centering
\renewcommand{\arraystretch}{1.0}
 \caption{\lia{Accuracies of different methods on Massachusetts dataset (1 m/pixel). (\%)}}
\begin{threeparttable}\resizebox{\textwidth}{!}{
\begin{tabular}{P{5cm}|P{1.9cm}|P{1.9cm}|P{1.9cm}|P{1.9cm}|P{1.9cm}|P{1.9cm}}
\Xhline{3\arrayrulewidth}

   \hline\hline
       \multirow{3}{*}{Method} & \multicolumn{2}{c|}{labeled:unlabeled $\approx$ 1:2} & \multicolumn{2}{c|}{labeled:unlabeled $\approx$ 1:5} & \multicolumn{2}{c}{labeled:unlabeled $\approx$ 1:10}\\
      ~ & \multicolumn{2}{c|}{1100 labeled, 2324 unlabeled} & \multicolumn{2}{c|}{560 labeled, 2864 unlabeled} & \multicolumn{2}{c}{300 labeled, 3124 unlabeled} \\
      \cline{2-7}
      ~ & F1 score & IoU & F1 score & IoU & F1 score & IoU \\
      \hline
      \begin{bfseries}Proposed method\end{bfseries} &\begin{bfseries} \lic{70.26 $\pm$ 0.57} \end{bfseries}& \begin{bfseries} \lic{54.15 $\pm$ 0.68} \end{bfseries} &\begin{bfseries} 68.59 \end{bfseries}&\begin{bfseries} 52.20 \end{bfseries} &\begin{bfseries} 67.69 \end{bfseries} &\begin{bfseries} 51.16 \end{bfseries}\\
      \hline
      SL & \lic{66.31 $\pm$ 0.40} & \lic{49.65 $\pm$ 0.47} & 62.75 & 45.72 & 57.91 & 40.76\\
        \hline
      SL + DA & \lic{66.56 $\pm$ 0.44} & \lic{49.85 $\pm$ 0.52} & 63.26 & 46.26 & 58.76 & 41.60\\
      \hline
      ICT \cite{verma2019interpolation} & \lic{67.03 $\pm$ 0.19} & \lic{50.42 $\pm$ 0.23} & 63.33 & 46.34 & 60.19 & 43.05\\
      \hline
      VAT \cite{miyato2018virtual} & \lic{66.10 $\pm$ 0.70} & \lic{49.40 $\pm$ 0.84} & 64.45 & 47.55 & 60.77& 43.65\\
      \hline
      CutMix \cite{french2020semi} & \lic{66.84 $\pm$ 0.32} & \lic{50.22 $\pm$ 0.38} & 63.13 & 46.13 & 59.41 & 42.26\\
      \hline
      CCT \cite{ouali2020semi} & \lic{67.79 $\pm$ 0.33} & \lic{51.30 $\pm$ 0.39} & 64.54 & 47.64 & 60.70 & 43.58\\
      \hline
 \lia{CR \cite{wang2020semi}} & \lic{65.83 $\pm$ 0.55} & \lic{51.08 $\pm$ 0.66} & \lia{63.91} & \lia{46.96} & \lia{60.68} & \lia{43.56}\\  
      \hline
      \lia{PiCoCo \cite{kang2021picoco}} & \lic{68.76 $\pm$ 0.52} & \lic{52.39 $\pm$ 0.62} & \lia{65.73} & \lia{48.96} & \lia{64.76} & \lia{47.88}\\   
\Xhline{3\arrayrulewidth}
\end{tabular}}
\end{threeparttable}
 \label{Tab.s3}
 \end{table*}

\begin{table*}[!t]
\centering
\renewcommand{\arraystretch}{1.0}
 \caption{\lia{Accuracies of different methods on Inria dataset (0.3 m/pixel). (\%)}}
\begin{threeparttable}\resizebox{\textwidth}{!}{
\begin{tabular}{P{5cm}|P{1.9cm}|P{1.9cm}|P{1.9cm}|P{1.9cm}|P{1.9cm}|P{1.9cm}}
\Xhline{3\arrayrulewidth}

   \hline\hline
       \multirow{3}{*}{Method} & \multicolumn{2}{c|}{labeled:unlabeled $\approx$ 1:2} & \multicolumn{2}{c|}{labeled:unlabeled $\approx$ 1:5} & \multicolumn{2}{c}{labeled:unlabeled $\approx$ 1:10} \\
      ~ & \multicolumn{2}{c|}{13000 labeled, 26852 unlabeled} & \multicolumn{2}{c|}{6000 labeled, 33852 unlabeled} & \multicolumn{2}{c}{3600 labeled, 36252 unlabeled} \\
      \cline{2-7}
      ~ & F1 score & IoU & F1 score & IoU & F1 score & IoU \\
      \hline
      \begin{bfseries}Proposed method\end{bfseries} &\begin{bfseries} 85.86 \end{bfseries}& \begin{bfseries} 75.22 \end{bfseries} &\begin{bfseries} 84.65 \end{bfseries}&\begin{bfseries} 73.39 \end{bfseries} &\begin{bfseries} 83.74 \end{bfseries} &\begin{bfseries} 72.03 \end{bfseries}\\
      \hline
      SL & 81.94 & 69.41 & 80.38 & 67.61 & 77.87 & 64.12\\
      \hline
      SL + DA & 82.40 & 70.07 & 81.01 & 68.08 & 78.56 & 64.69\\
      \hline
      ICT \cite{verma2019interpolation} & 82.53 & 70.26 & 81.10 & 68.21 & 78.60 & 64.75\\
      \hline
      VAT \cite{miyato2018virtual} & 82.79 & 70.63 & 81.42 & 68.66  & 78.48 & 64.58\\
      \hline
      CutMix \cite{french2020semi} & 82.89 & 70.77 & 81.34 & 68.55  & 78.59 & 64.74\\
      \hline
      CCT \cite{ouali2020semi} & 85.21 & 74.23 & 83.74 & 72.02  & 83.00 & 70.93\\
      \hline
 \lia{CR \cite{wang2020semi}} & \lia{82.38} & \lia{70.04} & \lia{81.00} & \lia{68.05} & \lia{78.27} & \lia{64.30}\\  
      \hline
      \lia{PiCoCo \cite{kang2021picoco}} & \lia{84.59} & \lia{73.29} & \lia{83.65} & \lia{71.90} & \lia{80.91} & \lia{67.94}\\          
\Xhline{3\arrayrulewidth}
\end{tabular}}
\end{threeparttable}
 \label{Tab.s4}
 \end{table*}

\begin{figure*}[!t]
 \captionsetup[subfigure]{labelformat=empty}
 \begin{center}
  \subfloat[(a)]{\includegraphics[width=0.18\textwidth]{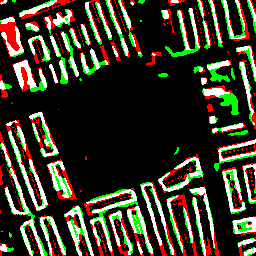}}
\hfil
   \subfloat[(b)]{\includegraphics[width=0.18\textwidth]{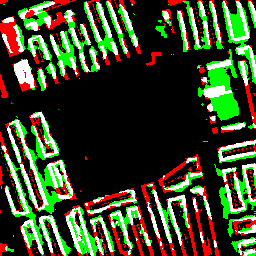}}
\hfil
   \subfloat[(c)]{\includegraphics[width=0.18\textwidth]{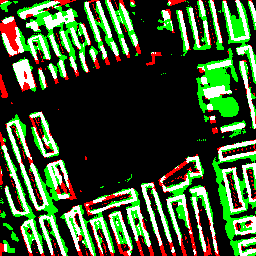}}
\hfil
    \subfloat[(d)]{\includegraphics[width=0.18\textwidth]{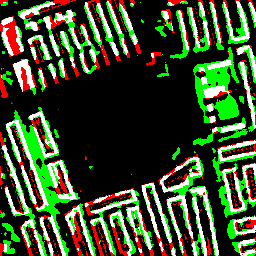}}
\hfil
  \subfloat[(e)]{\includegraphics[width=0.18\textwidth]{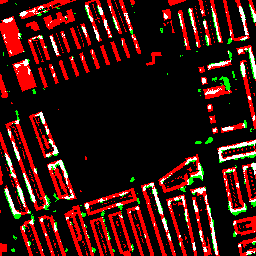}}
\vfil
   \subfloat[(f)]{\includegraphics[width=0.18\textwidth]{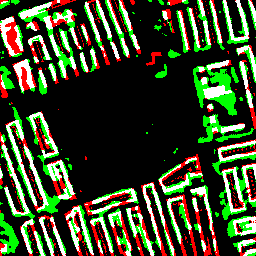}}
\hfil
   \subfloat[(g)]{\includegraphics[width=0.18\textwidth]{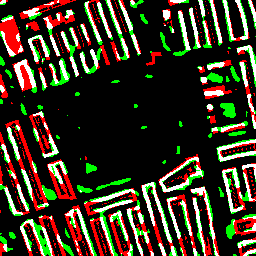}}
\hfil
   \subfloat[(h)]{\includegraphics[width=0.18\textwidth]{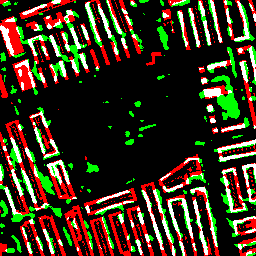}}
\hfil
   \subfloat[(i)]{\includegraphics[width=0.18\textwidth]{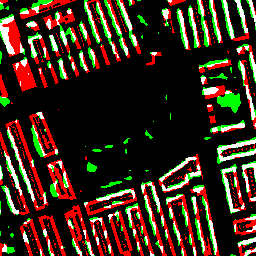}}
   \hfil
    \subfloat[(j)]{\includegraphics[width=0.18\textwidth]{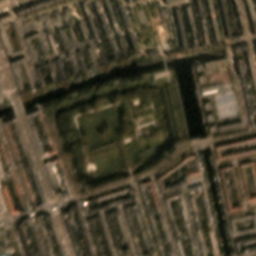}}

  \caption{\lia{Results obtained from  (a) proposed method, (b) SL, (c) SL+DA, (d) ICT \cite{verma2019interpolation}, (e) VAT \cite{miyato2018virtual}, (f) CutMix \cite{french2020semi}, (g) CCT \cite{ouali2020semi}, (h) CR \cite{wang2020semi}, and (i) PiCoCo \cite{kang2021picoco}.  In this experiment, the ratio of labeled data to unlabeled data is 1:10 (400 labeled, 4400 unlabeled). (j) is satellite imagery from the Planet dataset (spatial resolution: 3 m/pixel). Pixel-based true positives, false positives, and false negatives are marked in white, green, and red, respectively.}}
\label{Fig. 7}
\end{center}
 
\end{figure*}
 
\begin{figure*}[!t]
 \captionsetup[subfigure]{labelformat=empty}
 \begin{center}
  \subfloat[(a)]{\includegraphics[width=0.18\textwidth]{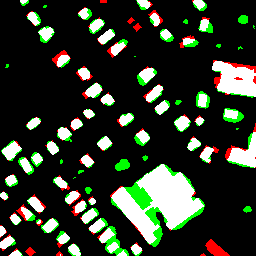}}
\hfil
   \subfloat[(b)]{\includegraphics[width=0.18\textwidth]{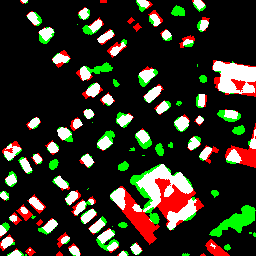}}
\hfil
   \subfloat[(c)]{\includegraphics[width=0.18\textwidth]{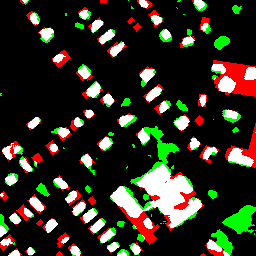}}
\hfil
    \subfloat[(d)]{\includegraphics[width=0.18\textwidth]{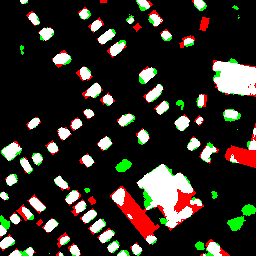}}
\hfil
  \subfloat[(e)]{\includegraphics[width=0.18\textwidth]{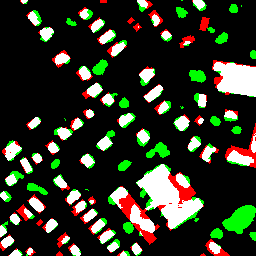}}
\vfil
   \subfloat[(f)]{\includegraphics[width=0.18\textwidth]{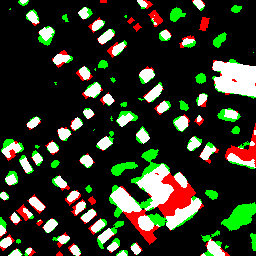}}
\hfil
   \subfloat[(g)]{\includegraphics[width=0.18\textwidth]{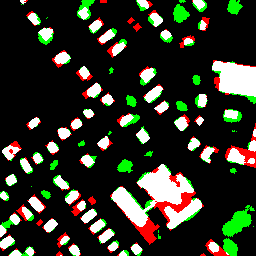}}
\hfil
   \subfloat[(h)]{\includegraphics[width=0.18\textwidth]{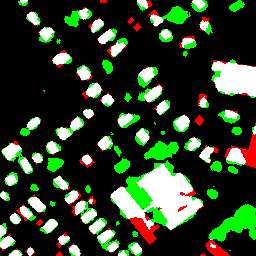}}
\hfil
   \subfloat[(i)]{\includegraphics[width=0.18\textwidth]{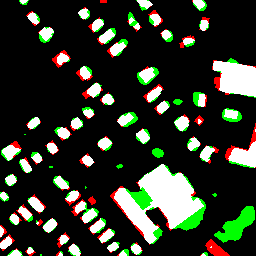}}
   \hfil
    \subfloat[(j)]{\includegraphics[width=0.18\textwidth]{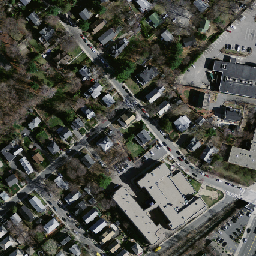}}

  \caption{\lia{Results obtained from  (a) proposed method, (b) SL, (c) SL+DA, (d) ICT \cite{verma2019interpolation}, (e) VAT \cite{miyato2018virtual}, (f) CutMix \cite{french2020semi}, (g) CCT \cite{ouali2020semi}, (h) CR \cite{wang2020semi}, and (i) PiCoCo \cite{kang2021picoco}.  In this experiment, the ratio of labeled data to unlabeled data is 1:10 (300 labeled, 3124 unlabeled). (j)  is aerial imagery from the Massachusetts dataset (spatial resolution: 1 m/pixel). Pixel-based true positives, false positives, and false negatives are marked in white, green, and red, respectively.}}
 \label{Fig. 8}
\end{center}

\end{figure*}

\begin{figure*}[!t]
 \captionsetup[subfigure]{labelformat=empty}
 \begin{center}
  \subfloat[(a)]{\includegraphics[width=0.18\textwidth]{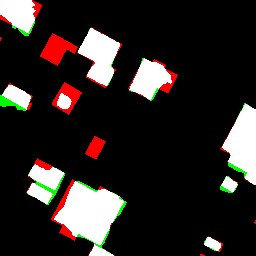}}
\hfil
   \subfloat[(b)]{\includegraphics[width=0.18\textwidth]{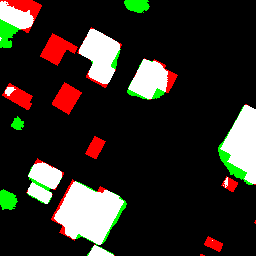}}
\hfil
   \subfloat[(c)]{\includegraphics[width=0.18\textwidth]{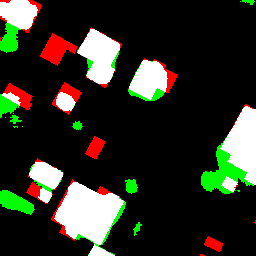}}
\hfil
    \subfloat[(d)]{\includegraphics[width=0.18\textwidth]{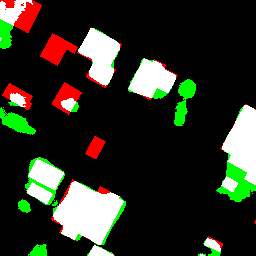}}
\hfil
  \subfloat[(e)]{\includegraphics[width=0.18\textwidth]{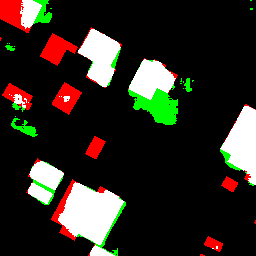}}
\vfil
   \subfloat[(f)]{\includegraphics[width=0.18\textwidth]{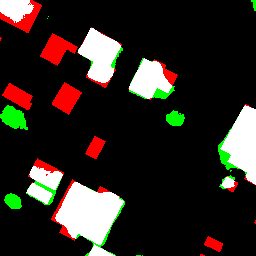}}
\hfil
   \subfloat[(g)]{\includegraphics[width=0.18\textwidth]{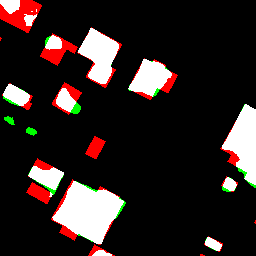}}
\hfil
   \subfloat[(h)]{\includegraphics[width=0.18\textwidth]{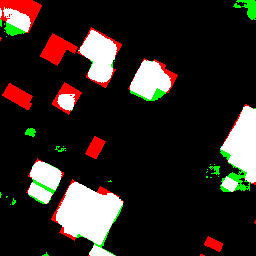}}
\hfil
   \subfloat[(i)]{\includegraphics[width=0.18\textwidth]{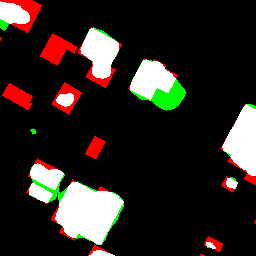}}
  \hfil
    \subfloat[(j)]{\includegraphics[width=0.18\textwidth]{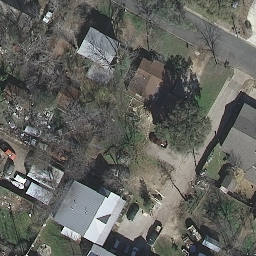}}

  \caption{\lia{Results obtained from  (a) proposed method, (b) SL, (c) SL+DA, (d) ICT \cite{verma2019interpolation}, (e) VAT \cite{miyato2018virtual}, (f) CutMix \cite{french2020semi}, (g) CCT \cite{ouali2020semi}, (h) CR \cite{wang2020semi}, and (i) PiCoCo \cite{kang2021picoco}.  In this experiment, the ratio of labeled data to unlabeled data is 1:10 (3600 labeled, 36252 unlabeled). (j)  is aerial imagery from the Inria dataset (spatial resolution: 0.3 m/pixel). Pixel-based true positives, false positives, and false negatives are marked in white, green, and red, respectively.}}
   \label{Fig. 9}
\end{center}
\end{figure*}

\lia{Furthermore, we make comparisons among the proposed method, SL, SL+DA, ICT \cite{verma2019interpolation}, VAT \cite{miyato2018virtual}, CutMix \cite{french2020semi}, CCT \cite{ouali2020semi}, CR \cite{wang2020semi} and PiCoCo \cite{kang2021picoco}. Here, the ratios of labeled data to unlabeled data are designed as 1:2, 1:5, and 1:10, respectively. SL is regarded as the baseline method that is only trained with labeled data, while SL+DA is trained on the labeled data that are already augmented. Labeled and unlabeled data are jointly trained for the proposed method, ICT \cite{verma2019interpolation}, VAT \cite{miyato2018virtual}, CutMix \cite{french2020semi}, CCT \cite{ouali2020semi}, CR \cite{wang2020semi} and PiCoCo \cite{kang2021picoco}. Their performance is evaluated from quantitative (cf. Tables \ref{Tab.s2}, \ref{Tab.s3}, and \ref{Tab.s4}) perspectives. \lic{As an example, experiments are carried out for five runs on the Massachusetts dataset where the ratio of labeled data to unlabeled data is 1:2. This provides a fair comparison, and the corresponding F1 score and IoU are shown as mean and variance.} Fig. \ref{Fig. 7}, \ref{Fig. 8}, and \ref{Fig. 9} illustrate visual results obtained by different methods for the ratio 1:10.}

It can be seen from the statistics of three datasets that the proposed approach significantly boosts performance in F1 score and IoU when compared with other methods. The challenge induced by the ratio of 1:10 is the limited data representation for buildings, however, we notice that the proposed method still manages to perform better on three datasets when compared to its competitors. Our method gains improvements of 5.18\%, 10.40\%, 7.91\% in IoU than SL for the Planet, Massachusetts, and Inria datasets, respectively. In particular, on the Massachusetts dataset, the IoU of the proposed approach is improved by more than 7\%  when compared to other methods. \lia{When the ratio of labeled data to unlabeled data is 1:2, the number of labeled samples is already sufficient for SL, but our method still provides advantages over it. Note that the proposed approach performs even better than the other semantic segmentation networks (cf. Table \ref{Tab.s1}) that are trained on the full labeled sets. This proves that the effectiveness and robustness of the proposed approach for the task of building footprint generation.}

The accuracy metric of IoU obtained by our method for the ratio of 1:2 is higher than that for the ratio of 1:10. This suggests that using more labeled samples increases the overall performances (42.20\% vs. 36.78\% in the Planet dataset, \lic{54.15 $\pm$ 0.68 \% vs. 51.16\% in the Massachusetts dataset}, 75.22\% vs. 72.03\% in the Inria dataset). It should be mentioned that the proposed approach is capable of reducing the gap between the different ratios. For instance, Table \ref{Tab.s3} shows that the IoU produced by our method, which is trained on the data of ratio of 1:10, only drops 1\% than that of ratio of 1:5. This demonstrates that our method can obtain reliable segmentation results even when there is only a small number of annotated samples.

\lia{The visual results on the Planet dataset are illustrated in Fig. \ref{Fig. 7}. There is a lot of missed detection in results provided by SL, VAT \cite{miyato2018virtual}, CCT \cite{ouali2020semi}, CR \cite{wang2020semi} and PiCoCo \cite{kang2021picoco}}, as the number of labeled samples is insufficient. On the contrary, our method can extract more building structures. Fig. \ref{Fig. 9} presents results on the Inria dataset. It can be clearly seen that our method is able to avoid more false alarms than its competitors. This suggests that the proposed method has a better capability of utilizing unlabeled data to improve network performance.

\begin{figure}[!t]
  \captionsetup[subfigure]{labelformat=empty}
 \begin{center}
 \includegraphics[width=1.0\linewidth]{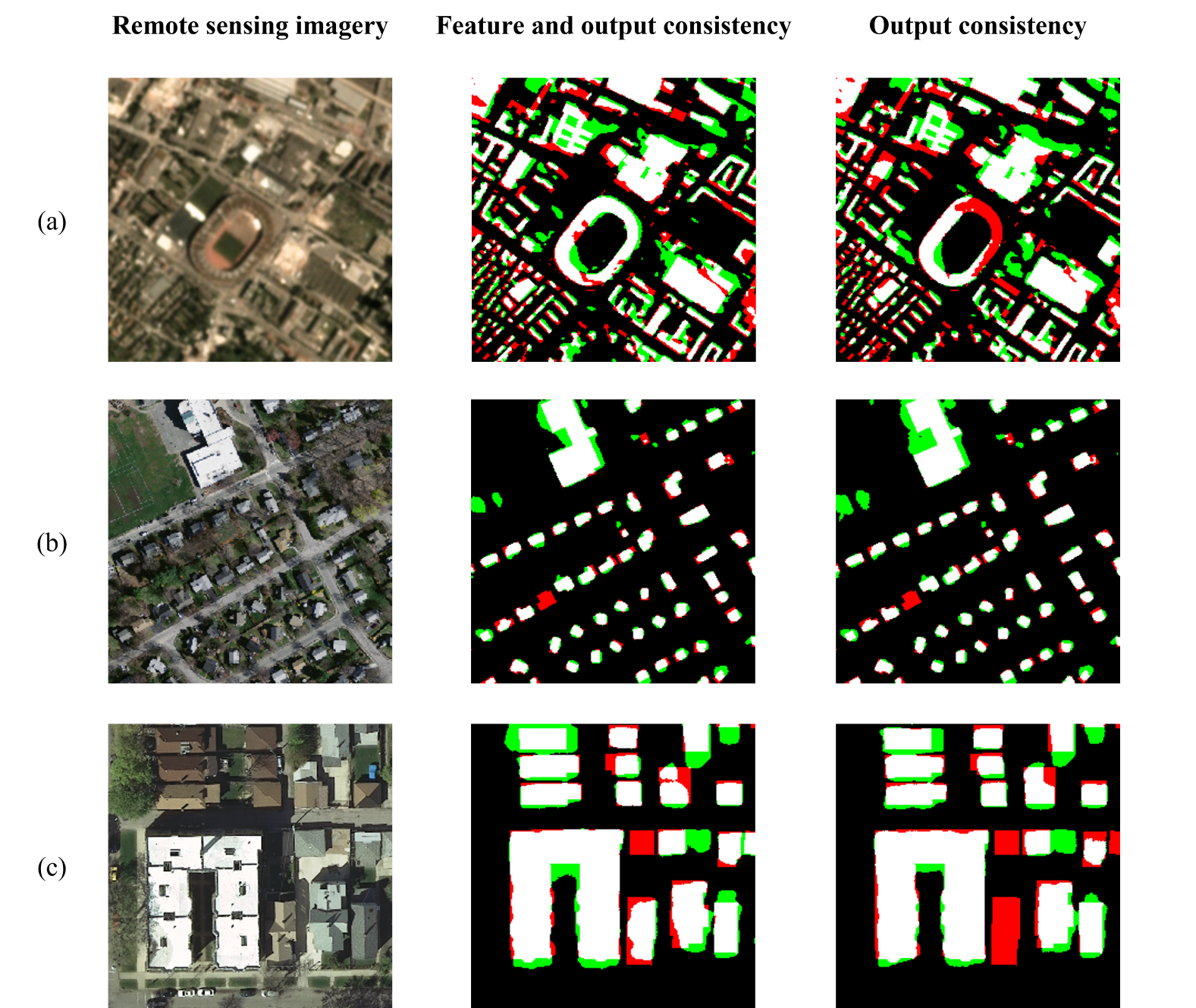}
  \caption{\lia{Results obtained from different methods on (a) Planet dataset (spatial resolution: 3 m/pixel),(b) Massachusetts dataset (spatial resolution: 1 m/pixel), and (c) Inria dataset (spatial resolution: 0.3 m/pixel). Pixel-based true positives, false positives, and false negatives are marked in white, green, and red, respectively.}}
  \label{Fig. 10}
  \end{center}
\end{figure}

\begin{table*}[!t]
\centering
\renewcommand{\arraystretch}{1.0}
 \caption{Ablation Study of the Imposed Consistency on Three datasets. (\%)}
\begin{threeparttable}\resizebox{\textwidth}{!}{
\begin{tabular}{P{5cm}|P{1.9cm}|P{1.9cm}|P{1.9cm}|P{1.9cm}|P{1.9cm}|P{1.9cm}}
\Xhline{3\arrayrulewidth}

   \hline\hline
      \multirow{3}{*}{The type of the imposed consistency} & \multicolumn{2}{c|}{Planet dataset (3 m/pixel)} & \multicolumn{2}{c|}{Massachusetts dataset (1 m/pixel)} & \multicolumn{2}{c}{Inria dataset (0.3 m/pixel)} \\
       ~ & \multicolumn{2}{c|}{1600 labeled, 3200 unlabeled} & \multicolumn{2}{c|}{1100 labeled, 2324 unlabeled} & \multicolumn{2}{c}{13000 labeled, 26852 unlabeled} \\
      \cline{2-7}
      ~ & F1 score & IoU & F1 score & IoU & F1 score & IoU \\
      \hline
      \begin{bfseries}Feature and output consistency\end{bfseries} &\begin{bfseries} 59.35 \end{bfseries}& \begin{bfseries} 42.20 \end{bfseries} &\begin{bfseries} 70.26 \end{bfseries}&\begin{bfseries} 54.16 \end{bfseries} &\begin{bfseries} 85.86 \end{bfseries} &\begin{bfseries} 75.22 \end{bfseries}\\  
      \hline
      Output consistency & 57.95 & 40.80 & 69.15 & 52.84 & 85.21 & 74.23\\
 
\Xhline{3\arrayrulewidth}
\end{tabular}}
\end{threeparttable}
 \label{Tab.s6}
 \end{table*}

\section{Discussion}
\label{sec:dis}
As shown in the results on three datasets for a semi-supervised setting, our proposed method with the ratio of 2:1 can deliver the best results. Therefore, in this section, we carry out ablation studies of the proposed method under this data split.

\subsection{Ablation Study of the Imposed Consistency}

\begin{figure*}[!t]
 \captionsetup[subfigure]{labelformat=empty}
 \begin{center}
  \subfloat[(a)]{\includegraphics[width=0.5\textwidth]{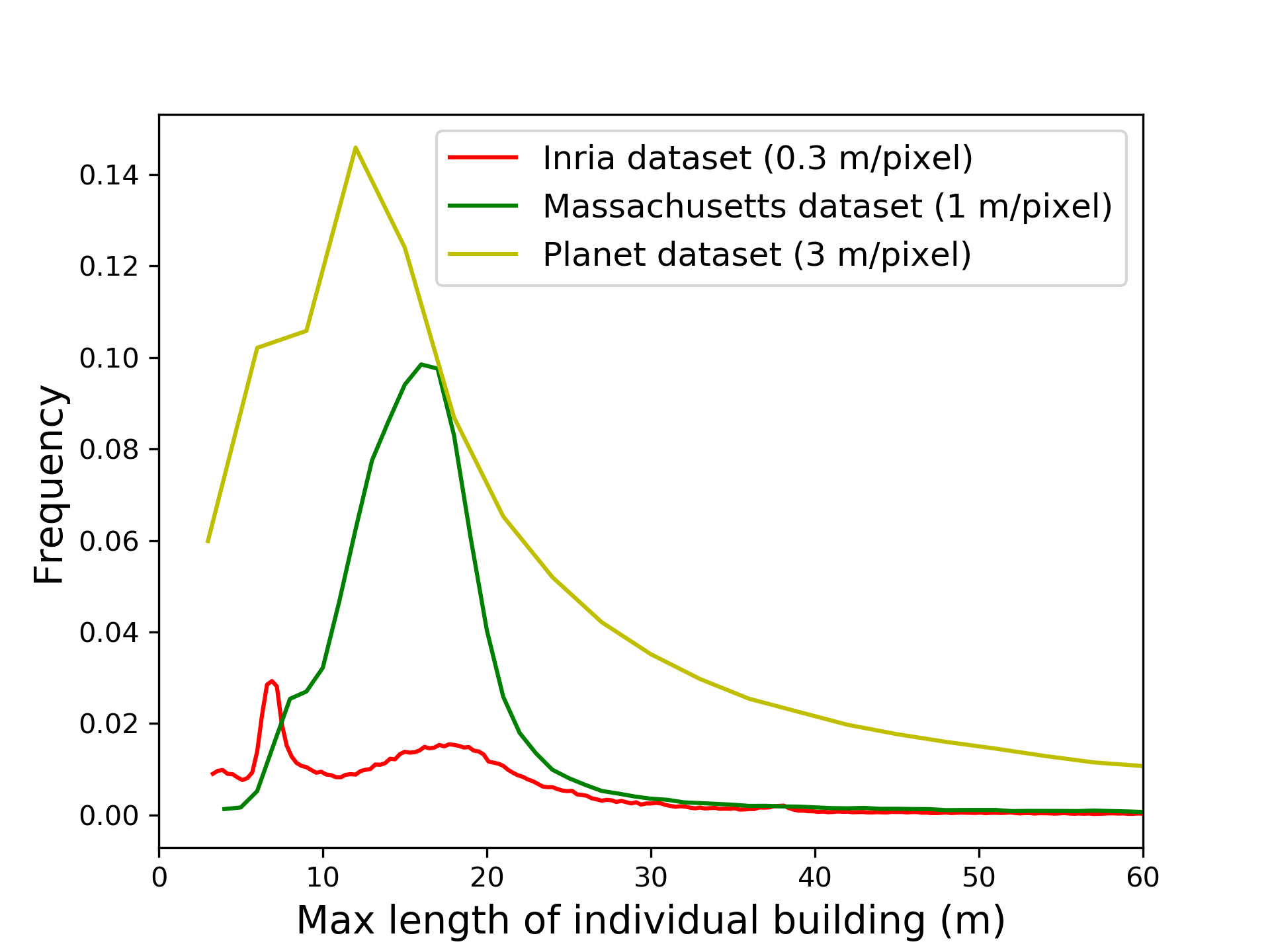}}
\hfil
   \subfloat[(b)]{\includegraphics[width=0.5\textwidth]{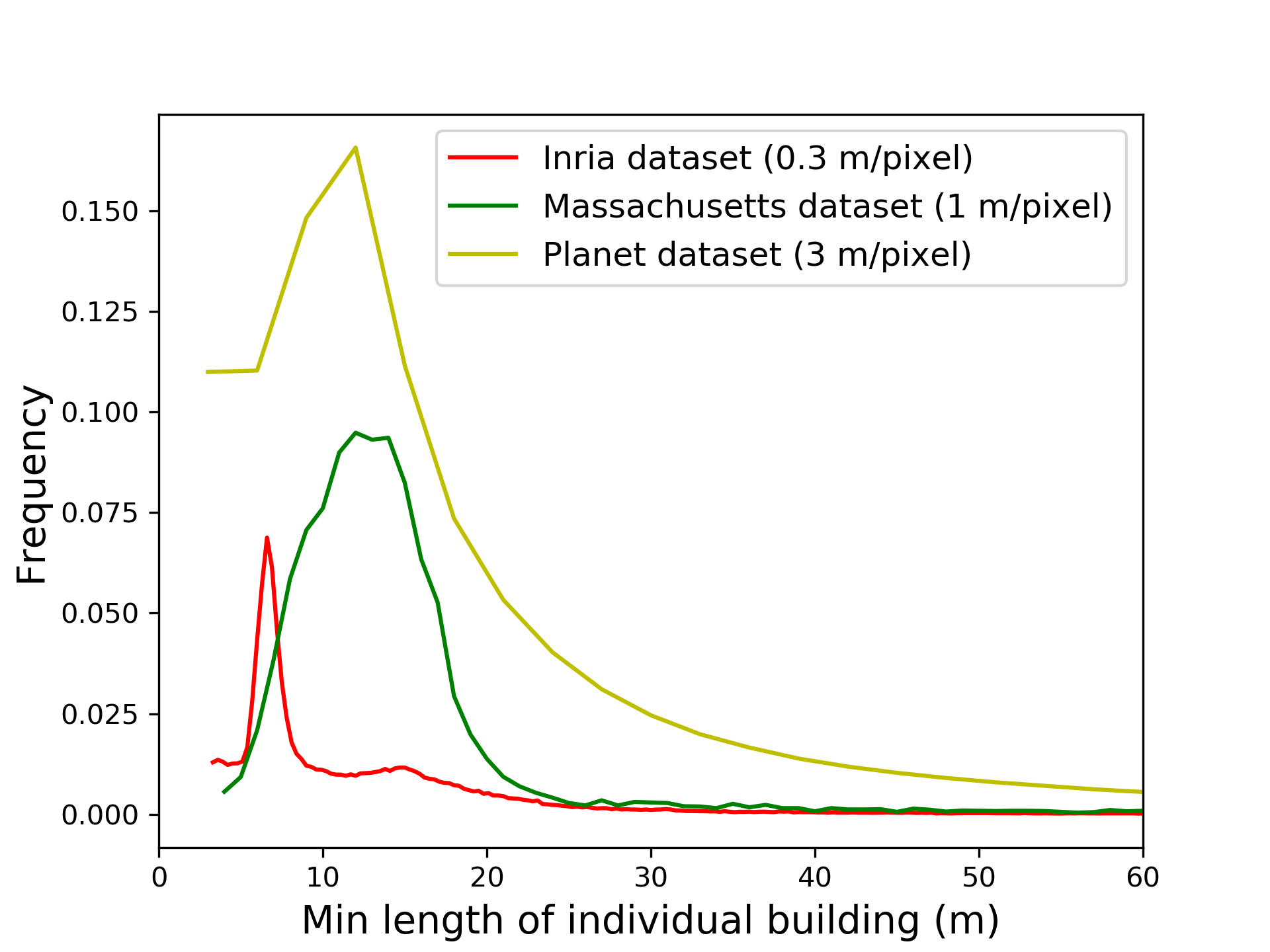}}
  \caption{\lia{Summarized statistics of (a) max length and (b) min length of individual buildings on three datasets.}}
  \label{Fig. a}
  \end{center}
\end{figure*}

\begin{table*}[!t]
\centering
\renewcommand{\arraystretch}{1.0}
 \caption{Ablation Study of the Assigned Perturbation on three datasets. (\%)}
\begin{threeparttable}\resizebox{\textwidth}{!}{
\begin{tabular}{P{5cm}|P{1.9cm}|P{1.9cm}|P{1.9cm}|P{1.9cm}|P{1.9cm}|P{1.9cm}}
\Xhline{3\arrayrulewidth}

   \hline\hline
      \multirow{3}{*}{The position of the assigned perturbation} & \multicolumn{2}{c|}{Planet dataset (3 m/pixel)} & \multicolumn{2}{c|}{Massachusetts dataset (1 m/pixel)} & \multicolumn{2}{c}{Inria dataset (0.3 m/pixel)} \\
       ~ & \multicolumn{2}{c|}{1600 labeled, 3200 unlabeled} & \multicolumn{2}{c|}{1100 labeled, 2324 unlabeled} & \multicolumn{2}{c}{13000 labeled, 26852 unlabeled} \\
      \cline{2-7}
      ~ & F1 score & IoU & F1 score & IoU & F1 score & IoU \\
      \hline
      d=1 & 57.72 & 40.47 & 67.67 & 51.46 & 84.65 & 73.59\\
      \hline
      d=2 &\begin{bfseries} 59.35 \end{bfseries}& \begin{bfseries} 42.20 \end{bfseries} & 68.66 & 52.27 & 84.44 & 73.28\\   
      \hline
      d=3 & 57.32 & 40.18 & 67.02 & 51.08 & 84.74 & 73.73\\   
      \hline
      d=4 & 56.58 & 39.45 &\begin{bfseries} 70.26 \end{bfseries}&\begin{bfseries} 54.16\end{bfseries} & 84.67 & 73.63\\   
      \hline      
      d=5 & 59.63 & 39.50 & 67.65 & 51.11 &\begin{bfseries} 85.86 \end{bfseries}&\begin{bfseries} 75.22 \end{bfseries}\\   
\Xhline{3\arrayrulewidth}
\end{tabular}}
\end{threeparttable}
 \label{Tab.s5}
 \end{table*}

\begin{figure}[!t]
  \captionsetup[subfigure]{labelformat=empty}
 \begin{center}
  \includegraphics[width=1.0\linewidth]{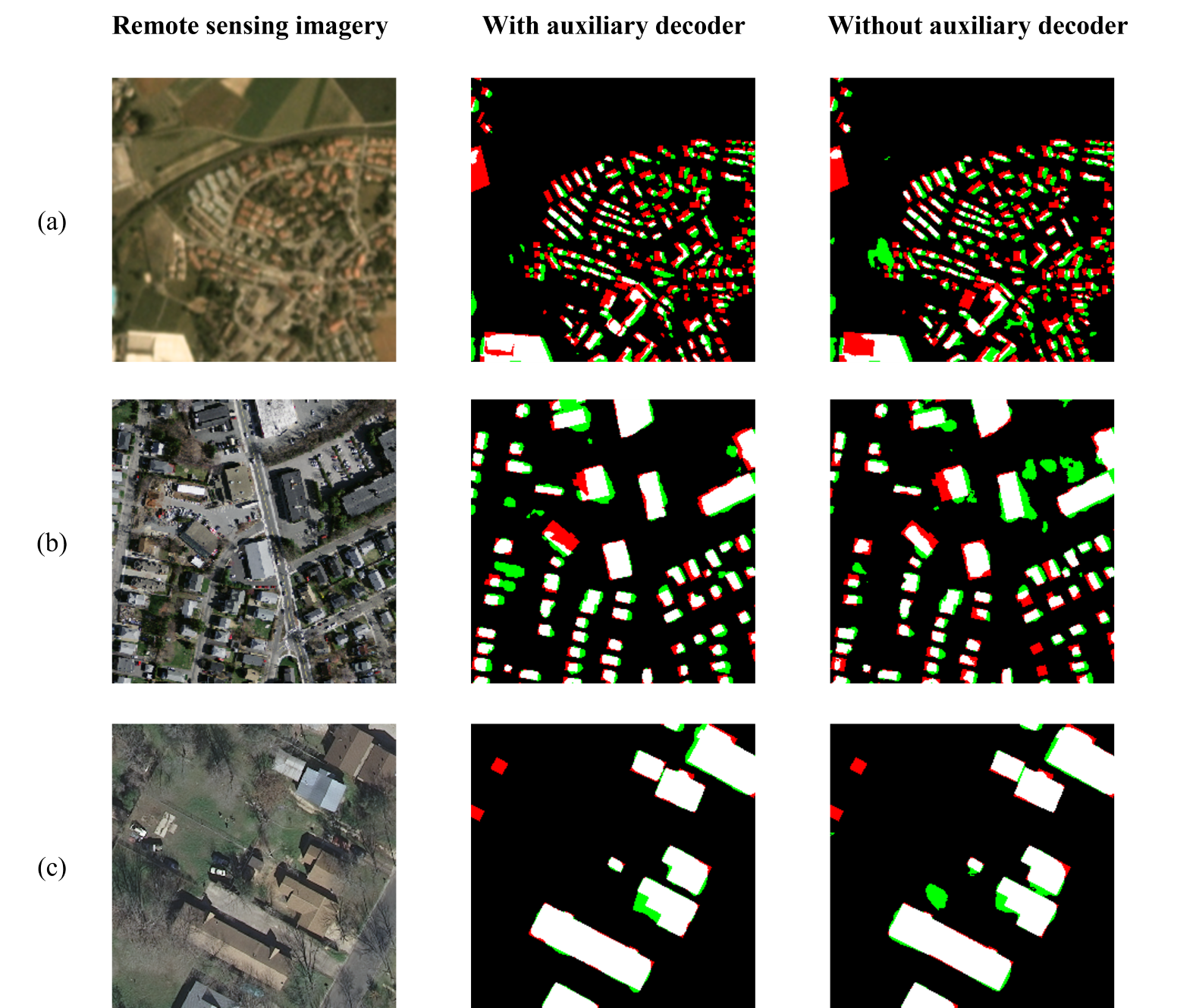}
  \caption{\lia{Results obtained from different methods on (a) Planet dataset (spatial resolution: 3 m/pixel),(b) Massachusetts dataset (spatial resolution: 1 m/pixel), and (c) Inria dataset (spatial resolution: 0.3 m/pixel). Pixel-based true positives, false positives, and false negatives are marked in white, green, and red, respectively.}}
  \label{Fig. 10a}
  \end{center}
\end{figure}

\begin{table*}[!t]
\centering
\renewcommand{\arraystretch}{1.0}
 \caption{\lia{Ablation Study of the Auxiliary Decoder on Three datasets. (\%)}}
\begin{threeparttable}\resizebox{\textwidth}{!}{
\begin{tabular}{P{5cm}|P{1.9cm}|P{1.9cm}|P{1.9cm}|P{1.9cm}|P{1.9cm}|P{1.9cm}}
\Xhline{3\arrayrulewidth}

   \hline\hline
      \multirow{3}{*}{Method} & \multicolumn{2}{c|}{Planet dataset (3 m/pixel)} & \multicolumn{2}{c|}{Massachusetts dataset (1 m/pixel)} & \multicolumn{2}{c}{Inria dataset (0.3 m/pixel)} \\
       ~ & \multicolumn{2}{c|}{1600 labeled, 3200 unlabeled} & \multicolumn{2}{c|}{1100 labeled, 2324 unlabeled} & \multicolumn{2}{c}{13000 labeled, 26852 unlabeled} \\
      \cline{2-7}
      ~ & F1 score & IoU & F1 score & IoU & F1 score & IoU \\
      \hline
      \begin{bfseries} With auxiliary decoder\end{bfseries} &\begin{bfseries} 59.35 \end{bfseries}& \begin{bfseries} 42.20 \end{bfseries} &\begin{bfseries} 70.26 \end{bfseries}&\begin{bfseries} 54.16 \end{bfseries} &\begin{bfseries} 85.86 \end{bfseries} &\begin{bfseries} 75.22 \end{bfseries}\\  
      \hline
      Without auxiliary decoder & 58.37 & 41.20 & 68.73 & 52.34 & 84.84 & 73.67\\
 
\Xhline{3\arrayrulewidth}
\end{tabular}}
\end{threeparttable}
 \label{Tab.s6a}
 \end{table*}

One contribution of our approach worthy of being highlighted is that we introduce a novel objective function by imposing consistency on both features and outputs between the main decoder and the auxiliary decoder.

The statistical results of different types of the imposed consistency are reported in Table \ref{Tab.s6}. Experimental results show that implementing feature and output consistency for this task is helpful to improve the network performance, and we can see nearly 1\% gains in IoU on all datasets when compared to solely output consistency. This may be because that more abstract and invariant information are included in the feature representations \cite{bengio2013representation}, and the network is able to learn more knowledge when feature consistency is additionally imposed. 

Fig. \ref{Fig. 10} illustrates a visual comparison between different types of the imposed consistency. Some buildings are omitted in the results provided by sole output consistency in the example areas of the INRIA dataset. The reason is that the sole output consistency ignores the rich information in feature representations. On the contrary, building masks obtained by the feature and output consistency are much closer to real building shapes. This suggests that our method can capture information in both feature representations and outputs, enabling the enhancement of semantic information of buildings.

\subsection{Ablation Study of the Assigned Perturbation}

For the perturbation being assigned to the feature representations within the encoder, we propose an instruction to select the optimal position: the encoder depth $d$. To verify this instruction, we apply the perturbation to five different positions within the encoder, respectively. Specifically, $d$ is first set as five numbers i.e., 1, 2, 3, 4, and 5, to investigate its impact on final results. The spatial size of their corresponding feature maps is $128 \times 128$, $64 \times 64$, $32 \times 32$, $16 \times 16$, $8 \times 8$. 

The statistical results of the perturbation applied to different depths within the encoder are shown in Table \ref{Tab.s5}. We can see that the best position to assign the perturbation is varied across different datasets. Moreover, increasing the value of the depth will promote the improvement of results on the higher resolution dataset (Inria dataset). However, we note that a large value of $d$ will lead to a reduction in accuracy metrics on the relatively low-resolution dataset (Planet dataset). The best results are obtained when $d=2$ for the Planet dataset, $d=4$ for the Massachusetts dataset, and $d=5$ for the Inria dataset. This coincides with our proposed instruction to apply the perturbation. 

Taking the spatial resolution of remote sensing imagery into consideration, the respective field of these positions are corresponding to  $3 \times 2^2= 12 m$  (Planet dataset), $ 1 \times 2^4 = 16 m$ (Massachusetts dataset), $0.3 \times 2^5 = 9.6 m$  (Inria dataset), which are close to the size of a building that usually has a length within the range from 10 m to 20 m.  Afterward, we calculate the statistics of individual buildings of all three datasets, i.e., max length and min length (cf. Fig. \ref{Fig. a}). We found that the mean values of the max length of individual buildings are 19 m for the Planet dataset, 17 m for the Massachusetts dataset, 16 m for the Inria dataset. Mean values of the min length of individual buildings are 17 m for the Planet dataset, 14 m for the Massachusetts dataset, 12 m for the Inria dataset. That is to say, mean values of max length and min length of individual buildings also range from 10 m to 20 m among all datasets. This indicates the geometrical characteristics of the building are related to the effective receptive field of the network, which may place an emphasis on how to select the optimal position to assign the perturbation in the whole framework. Therefore, we infer that the perturbation should be assigned to the different positions within the encoder according to the spatial resolution of remote sensing imagery and the mean size of the individual buildings in the study area.

\subsection{\lia{Ablation Study of the Auxiliary Decoder}}

\lia{In our approach, an auxiliary decoder is employed to train the unlabeled set, and additional training signals can be extracted by enforcing the consistency of features and predictions between the main decoder and the auxiliary decoders. In order to validate the effectiveness of the auxiliary decoder, we perform an ablation study with another competitor, i.e., the proposed method without auxiliary decoder. That is to say, the auxiliary decoder is removed, and the main decoder takes as input both an uncorrupted and perturbed version of the encoder’s output to impose consistency on their features and outputs.}

\lia{The ablation study is carried out on Planet, Massachusetts, and Inria datasets. Numerical results are shown in Tables \ref{Tab.s6a}. As can be seen in statistical results on all three datasets, an auxiliary decoder brings a nearly 1\% improvement in IoU, leading to a positive influence on the performance of our network. Fig. \ref{Fig. 10a} shows a visual comparison of segmentation results, which demonstrates that the performance of our approach can be boosted up by the leverage of an auxiliary decoder. In Fig. \ref{Fig. 10a} (e) and (h), the method without auxiliary decoder wrongly identifies cars as buildings on both Massachusetts and Inria datasets. This is because, the colors of cars are similar to those of buildings, which leads to a misjudgment. The use of an auxiliary decoder is able to avoid such false alarms. The main reason is that supervision from the same decoder might guide the network to better approximate the features and outputs of the perturbed inputs, making the network converges in the wrong direction. In contrast, supervision by the features and predictions from the other decoder is able to avoid over-fitting the wrong direction.}

\section{Conclusion}
\label{sec:con}
Considering that the performance of semantic segmentation networks is limited when the annotated training samples are insufficient, a novel semi-supervised building footprint generation method with feature and output consistency training is proposed in this paper. The proposed model comprises three modules: a shared encoder, a main decoder, and an auxiliary decoder. More specifically, the shared encoder and the main decoder are designed to learn from labeled data in a fully supervised manner. Afterward, we assign the perturbation at the intermediate feature representations within the encoder and aims to encourage the auxiliary decoder to give consistent predictions for unlabeled inputs as the main decoder. The consistency is imposed between outputs and features of the main decoder and those of the auxiliary decoder. The performance of the proposed end-to-end network is assessed on three datasets with different resolutions: Planet dataset (3 m/pixel), Massachusetts dataset (1 m/pixel), and Inria dataset (0.3 m/pixel).  Experimental results suggest that the incorporation of both feature and output consistency in our method can offer more satisfactory building footprints, where omission errors can be alleviated to a large extent. Therefore, We believe that our method is a robust solution for building footprint generation when dealing with scarce training samples. Furthermore, the best position to assign the perturbation has been investigated that the perturbation should be applied to the different depths within the encoder according to the spatial resolution of input remote sensing imagery and the mean size of the individual buildings in the study area. This practical strategy is beneficial to other semi-supervised building footprint generation works that use remote sensing imagery. A subsequent study will intend to investigate the potential of the feature and output consistency training in the instance segmentation of buildings.

\bibliography{Reference}
\bibliographystyle{IEEEtran}
\begin{IEEEbiography}
[{\includegraphics[width=1in,height=1.25in,clip,keepaspectratio]{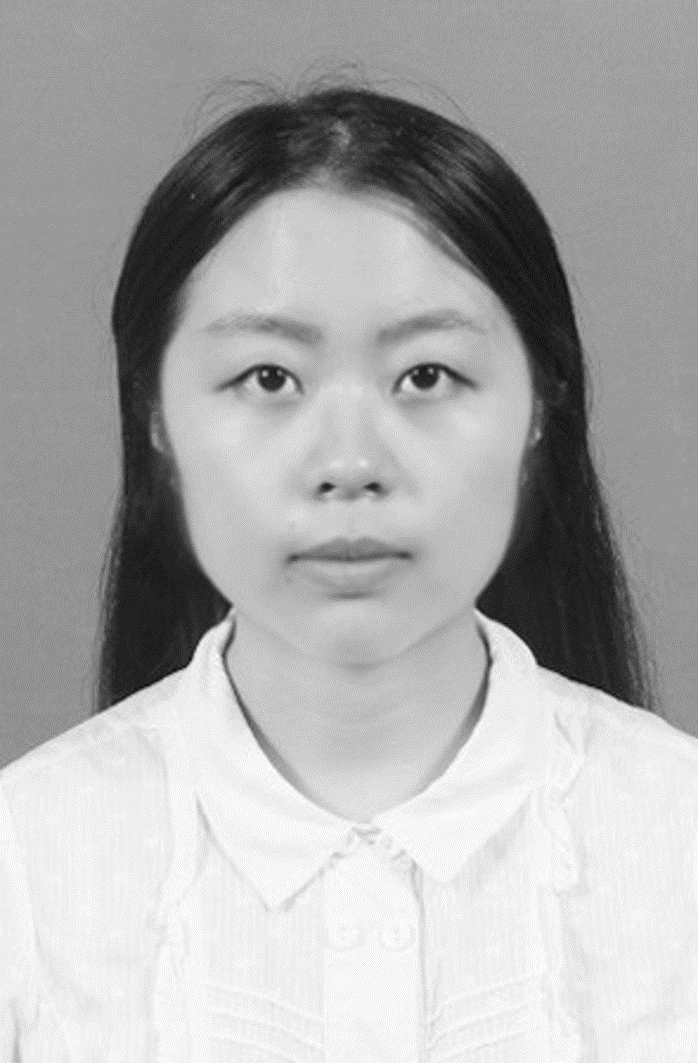}}] {Qingyu Li}
(S'21) received the bachelor's degree in remote sensing science and technology from
the Wuhan University, Wuhan, China, in 2015, and
the master's degree in Earth Oriented Space Science and Technology (ESPACE) from the Technische Universit{\"a}t M{\"u}nchen (TUM), Munich, Germany, in
2018. 

She is currently pursuing the Ph.D. degree with the TUM, Munich, Germany and the German Aerospace Center (DLR), Wessling, Germany. Her research interests include deep learning, remote sensing mapping, and remote sensing applications.
\end{IEEEbiography}

\begin{IEEEbiography}
[{\includegraphics[width=1in,height=1.25in,clip,keepaspectratio]{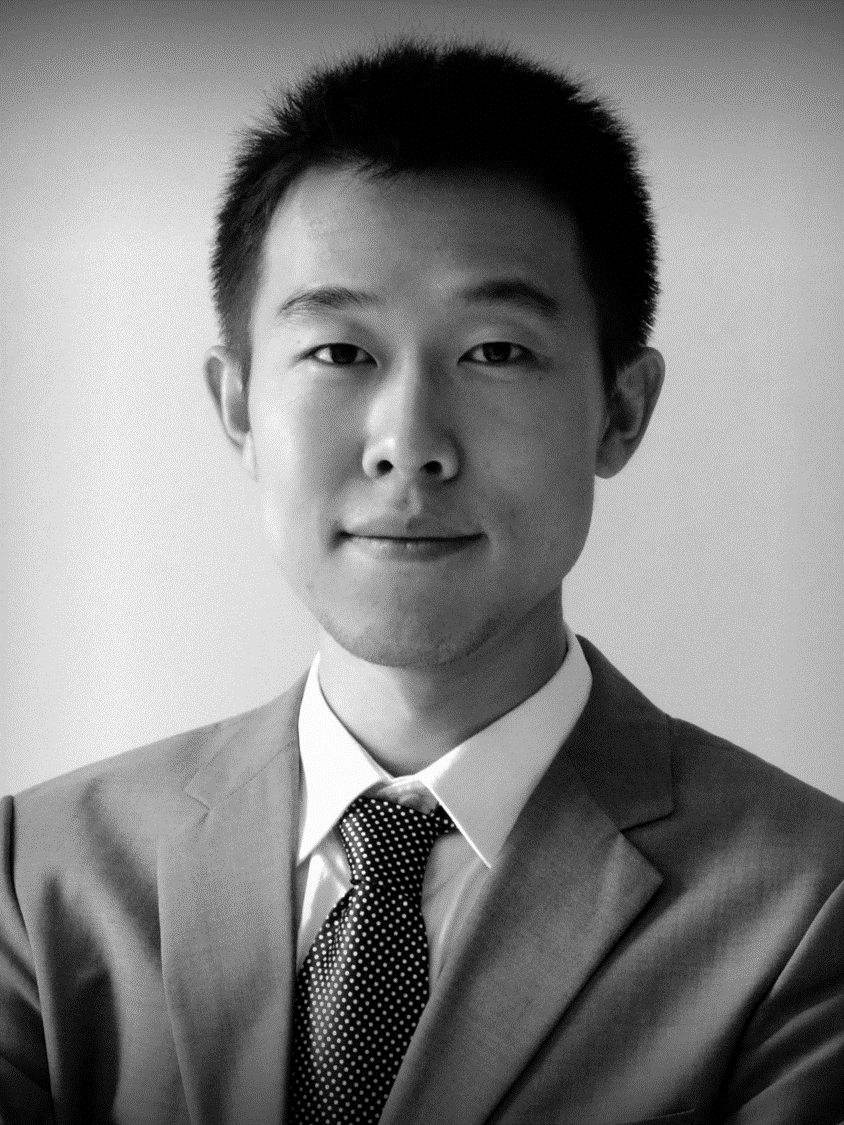}}] {Yilei Shi}
(M'18) received the Dipl.-Ing degree in mechanical engineering and Dr.-Ing degree in signal processing from the Technische Universit{\"a}t M{\"u}nchen (TUM), Munich, Germany, in 2010 and 2019, respectively. In April and May 2019, he was a guest scientist with the department of applied mathematics and theoretical physics, University of Cambridge, United Kingdom. He is currently a senior scientist with the Chair of Remote Sensing Technology, TUM.

His research interests include fast solver and parallel computing for large-scale problems, high performance computing and computational intelligence, advanced methods on SAR and InSAR processing, machine learning and deep learning for variety of data sources, such as SAR, optical images, and medical images, and PDE-related numerical modeling and computing.
\end{IEEEbiography}

\begin{IEEEbiography}[{\includegraphics[width=1in,height=1.25in,clip,keepaspectratio]{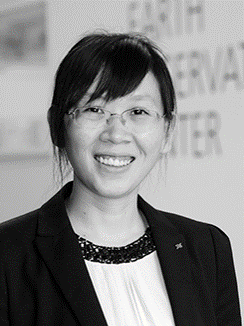}}]{Xiao Xiang Zhu}(S'10--M'12--SM'14--F'21) received the Master (M.Sc.) degree, her doctor of engineering (Dr.-Ing.) degree and her “Habilitation” in the field of signal processing from Technical University of Munich (TUM), Munich, Germany, in 2008, 2011 and 2013, respectively.
\par
She is currently the Professor for Data Science in Earth Observation (former: Signal Processing in Earth Observation) at Technical University of Munich (TUM) and the Head of the Department ``EO Data Science'' at the Remote Sensing Technology Institute, German Aerospace Center (DLR). Since 2019, Zhu is a co-coordinator of the Munich Data Science Research School (www.mu-ds.de). Since 2019 She also heads the Helmholtz Artificial Intelligence -- Research Field ``Aeronautics, Space and Transport". Since May 2020, she is the director of the international future AI lab "AI4EO -- Artificial Intelligence for Earth Observation: Reasoning, Uncertainties, Ethics and Beyond", Munich, Germany. Since October 2020, she also serves as a co-director of the Munich Data Science Institute (MDSI), TUM. Prof. Zhu was a guest scientist or visiting professor at the Italian National Research Council (CNR-IREA), Naples, Italy, Fudan University, Shanghai, China, the University  of Tokyo, Tokyo, Japan and University of California, Los Angeles, United States in 2009, 2014, 2015 and 2016, respectively. She is currently a visiting AI professor at ESA's Phi-lab. Her main research interests are remote sensing and Earth observation, signal processing, machine learning and data science, with their applications in tackling societal grand challenges, e.g. Global Urbanization, UN’s SDGs and Climate Change.

Dr. Zhu is a member of young academy (Junge Akademie/Junges Kolleg) at the Berlin-Brandenburg Academy of Sciences and Humanities and the German National  Academy of Sciences Leopoldina and the Bavarian Academy of Sciences and Humanities. She serves in the scientific advisory board in several research organizations, among others the German Research Center for Geosciences (GFZ) and Potsdam Institute for Climate Impact Research (PIK). She is an associate Editor of IEEE Transactions on Geoscience and Remote Sensing and serves as the area editor responsible for special issues of IEEE Signal Processing Magazine. She is a Fellow of IEEE.
\end{IEEEbiography}

\end{document}